\documentclass[final]{clv2025_arxiv}

\jvol{vv}
\jnum{nn}
\jyear{2025}

\dochead{} 

\pageonefooter{}

\usepackage{amsmath}
\DeclareMathOperator*{\argmin}{argmin}
\usepackage{booktabs}

\usepackage[dvipsnames]{xcolor}
\usepackage{arydshln}
\usepackage{multirow}
\usepackage[normalem]{ulem}
\usepackage{algorithm}
\usepackage{algpseudocode}
\newcommand{\COMMENT}[1]{}
\usepackage{enumitem}
\setlist[description]{leftmargin=\parindent,labelindent=\parindent}
\usepackage{array}

\runningtitle{Source-Aware MT Metrics for ST}
\runningauthor{Cettolo, Gaido, Negri, Papi, Bentivogli}

\begin{document}

\title{How to Evaluate Speech Translation with Source-Aware Neural MT Metrics}

\author{Mauro Cettolo\thanks{Corresponding author}, Marco Gaido,
Matteo Negri, \\ Sara Papi, Luisa Bentivogli}

\affilblock{
    \affil{Fondazione Bruno Kessler\\\quad \email{cettolo@fbk.eu}\\\quad
    \email{mgaido@fbk.eu}\\\quad
    \email{negri@fbk.eu}\\\quad
    \email{spapi@fbk.eu}\\\quad
    \email{bentivo@fbk.eu}}
}

\maketitle

\begin{abstract}

Automatic evaluation of speech-to-text translation (ST) systems is typically performed by comparing translation hypotheses with one or more reference translations. While effective to some extent, this approach inherits the limitation of reference-based evaluation that ignores valuable information from the source input. In machine translation (MT), recent progress has shown that neural metrics incorporating the source text achieve stronger correlation with human judgments. Extending this idea to ST, however, is not trivial because the source is audio rather than text, and reliable transcripts or alignments between source and references are often unavailable. In this work, we conduct the first systematic study of source-aware metrics for ST, with a particular focus on real-world operating conditions where source transcripts are not available. We explore two complementary strategies for generating textual proxies of the input audio, automatic speech recognition (ASR) transcripts, and back-translations of the reference translation, and introduce a novel two-step cross-lingual re-segmentation algorithm to address the alignment mismatch between synthetic sources and reference translations. Our experiments, carried out on two ST benchmarks covering 79 language pairs and six ST systems with diverse architectures and performance levels, show that ASR transcripts constitute a more reliable synthetic source than back-translations when word error rate is below 20\%, while back-translations always represent a computationally cheaper but still effective alternative. The robustness of these findings is further confirmed by experiments on a low-resource language pair (Bemba-English) and by a direct validation against human quality judgments. Furthermore, our cross-lingual re-segmentation algorithm enables robust use of source-aware MT metrics in ST evaluation, paving the way toward more accurate and principled evaluation methodologies for speech translation.
\end{abstract}

\section{Introduction}
\label{sec:intro}

Translating between natural languages is a complex task even for humans, due to intrinsic linguistic challenges such as ambiguity and polysemy, dependence on context, structural differences across languages (e.g., Subject-Verb-Object vs. Verb-Object-Subject orders), the presence of idiomatic and figurative expressions, and the influence of pragmatics~\cite{jm3}. Despite these intrinsic difficulties, human translation can be assumed to be error-free, especially for professionals, while the same cannot be said for translation performed automatically by machines, neither from text to text (machine translation, MT) nor from speech to text (ST). Therefore, for automatic translation, an additional challenge arises, that of evaluating its quality. 

The evaluation of the quality of translations generated by machines can be performed either manually or automatically. Manual evaluation involves the annotation of systems' outputs by human professionals. While it is considered the most reliable option, it is rarely employed due to its large cost and the consequent infeasibility of performing it at scale \cite{freitag-etal-2021-experts}. For this reason, research advancements mostly rely on automatic metrics \cite{Marie2021ScientificCOA}, the focus of this work.

Traditionally, automatic MT metrics rely on comparing a system output against one or more human reference translations assumed to represent the ``correct'' rendering of the source sentence to translate. This is the case of BLEU \cite{papineni-etal-2002-bleu}, the most widespread MT metric in the scientific community over the last two decades \cite{mathur-etal-2020-tangled}, which computes $n$-gram overlaps between the system output (hypothesis) and the reference(s). However, following evidence of the mismatch between rankings produced by BLEU and by human evaluations \citep{bojar-etal-2018-findings,barrault-etal-2019-findings}, the community has undertaken efforts to build more reliable metrics. While it has not yet been possible to define a ``perfect'' holistic metric, as demonstrated by the fact that specific shared tasks on evaluation metrics are still annually organized,\footnote{Since 2008, the conference on Machine Translation (WMT) has organized a shared task on MT automatic evaluation metrics. Over the years, it has become the reference on the topic and annually attracts a large part of the scientific community working on it. The link to the 2025 edition is: \url{https://www2.statmt.org/wmt25/mteval-subtask.html}} recent years have seen the rise of neural metrics that do not rely solely on the similarity between hypotheses and references, but also take the source text into account. The first and most widespread metric of this type is COMET~\cite{rei-etal-2020-comet}, which demonstrated its effectiveness from its first participation in the MT Metrics shared task~\cite{mathur-etal-2020-results} within the WMT 2020 conference. Since then, other source-aware metrics have been proposed showing steady advancements~\cite{freitag-etal-2023-results,freitag-etal-2024-llms}.

In the related field of ST, the community has traditionally relied on the same metrics, evaluation procedures, and insights coming from MT. In particular, reference-based evaluation remains the same across the two fields, as the only difference is the modality of the source, which, in ST, is audio rather than text. However, the input dissimilarity means that the more recent and reliable neural source-aware MT metrics cannot be employed in ST because they require a textual source, not audio. As a solution, two approaches are possible: (i) designing novel multimodal metrics capable of directly exploiting the audio, or (ii) relying on a textual proxy for the source speech. In this paper, we focus on the second approach, which has recently been adopted also in the annual evaluation campaign organized by the International Conference on Spoken Language Translation (IWSLT)~\citep{agostinelli-etal-2025-findings}.  
In the last few years, IWSLT evaluation has indeed shifted from source-independent string-based metrics like BLEU to the source-aware neural metrics like COMET as the official ranking criterion. In this shift, however, the organizers have not explicitly disclosed which source was used to feed COMET, nor have they provided any evidence supporting the reliability of their choice, leaving room for dedicated research to support the choice of best practices. To this aim, given the proven superiority of source-aware metrics, it is important to understand whether automatic source generation is a viable option when manual transcripts are not available, a typical real-world condition which remains underexplored. To fill this gap, the first research question we address in this paper is~(\textbf{RQ1}): Can we automatically derive the source text corresponding to the audio without compromising the reliability of the source-aware metrics?

To answer RQ1, we study two alternative ways to generate a synthetic textual source: the automatic transcription of the audio with an automatic speech recognition~(ASR) system and the back-translation (BT) of the reference translation with an MT system. This leads to our second research question (\textbf{RQ2}): Which of the two is the best method to automatically generate synthetic textual sources for ST evaluation? If neither method proves consistently superior, what factors should guide the choice between them in a given setting?

We answer RQ2 in two different scenarios. First, we study the simpler case, in which the reference translation is paired with the corresponding segment of source audio. This is typical in controlled evaluation settings based on segment-level assessments. Then, we move to the more challenging setting, in which the ST benchmark is made of long audio recordings with corresponding document-level reference translations, without sentence-level audio-textual alignments. In this second scenario, the ASR-based solution requires a cross-lingual re-segmentation to pair the generated transcripts with the reference translations. This problem constitutes our last research question (\textbf{RQ3}): When the synthetic source text is not aligned with the reference text, can we re-align it without affecting the quality of the resulting evaluation? To answer RQ3, we propose a novel two-stage, cross-lingual algorithm, XLR-Segmenter,\footnote{XLR-Segmenter, licensed under Apache Version 2.0, is available on GitHub (\url{https://github.com/hlt-mt/source-resegmenter}) and on PyPi (\texttt{pip install source\_resegmenter}).} which we validate not only against the back-translation alternative but also against the easier scenario in which the audio segmentation is given.

Our investigation and evaluation of the proposed solutions is carried out in a rich experimental setup that leverages ST benchmarks in which human-labelled transcripts and sentence-level alignments are available. To ensure the soundness and robustness of our findings, we consider:

\begin{itemize}
    
\item Two ST benchmarks covering 79 language pairs from different language families, for a total of over 120,000 sentences, and a low-resource language pair (Bemba-English), to ensure the generalizability of our findings;

\item Two representative source-aware evaluation metrics (COMET and MetricX), selected based on recent studies showing their high correlation with human judgments;

\item Six ST systems, evenly divided between cascaded and direct architectures, spanning a wide performance range;

\item A test set including human quality judgments, to validate the synthetic metrics directly against human assessments.

\end{itemize}

Our empirical results highlight that both synthetic sources serve as effective textual proxies for input audio, enabling the application of source-aware metrics in ST (RQ1). At the same time, we also observe the superior reliability of automatic transcription over back translation, provided that its word error rate remains below 20\% (RQ2). Finally, the proposed cross-lingual re-segmentation algorithm proves to support reliable evaluation, yielding only negligible degradation in the absence of audio-text alignments~(RQ3).

The paper starts by providing an overview of the recent literature that defines the scope of our investigation (Section~\ref{sec:previous}). It then describes the solutions we propose (Section~\ref{sec:methodology}) and provides the experimental setup (Section~\ref{sec:exp-setting}). The main experiments (Section~\ref{sec:results}) are divided into four blocks. The first two blocks (Sections~\ref{sec:exp-src-controlled} and~\ref{sec:synth-source-analysis}), validate and compare the synthetic sources under controlled conditions, where source audio and reference translations are manually aligned, to observe their performance without external influence. The third block (Section~\ref{sec:reseg-controlled}) assesses the source re-segmentation algorithm under controlled conditions as well, using reference transcripts instead of automatically generated ones. Finally, in the fourth block (Section~\ref{sec:realistic}), experiments are conducted in the most realistic scenario possible, to validate all the proposed solutions ``in the wild''. Each block is organized by first providing the experimental outline, then presenting the results, and finally listing key observations and takeaways. The two following sections complement the main evaluation: Section~\ref{sec:additional-exps_low-resource} extends it to a low-resource setting involving the Bemba-English language pair, while Section~\ref{sec:additional-exps_DA} validates the synthetic metrics directly against human quality judgments. A thorough discussion (Section~\ref{sec:discussion}) and the examination of the limitations of the work (Section~\ref{sec:limitations}) conclude the paper.

\section{Related Works}
\label{sec:previous}

\subsection{Metrics for Machine Translation}
The automatic evaluation of MT outputs has been a central topic of research since the early days of statistical approaches~\cite{brown-etal-1990-statistical}. The first evaluation strategies relied on surface-based comparisons between system outputs and human references, with metrics such as BLEU~\cite{papineni-etal-2002-bleu}, TER~\cite{snover-etal-2006-study}, and chrF~\cite{popovic-2015-chrf}. While computationally efficient, early n-gram- or character-based metrics often failed to capture semantic adequacy or tolerate syntactic variation. To address these shortcomings, improved solutions such as METEOR \cite{banerjee-lavie-2005-meteor} incorporated synonym matching and paraphrasing, moving beyond strict n-gram overlap. More recently, the advent of neural metrics has further transformed MT evaluation by modeling semantic similarity directly through contextual embeddings and learned representations. Leveraging pretrained language models, approaches such as BERTScore~\cite{BERTScore}, BLEURT~\cite{sellam-etal-2020-bleurt}, COMET~\cite{rei-etal-2020-comet}, and MetricX~\cite{juraska-etal-2023-metricx} have demonstrated substantially stronger correlations with human judgments. This dominance of learned metrics over traditional surface-level scores has been repeatedly confirmed in recent WMT General MT Shared Tasks~\cite{freitag-etal-2022-results,freitag-etal-2023-results}. Since 2008, WMT has hosted a dedicated Metrics Shared Task~\cite{ws-2008-statistical}, providing systematic comparisons of evaluation metrics across systems, languages, and evaluation conditions~\cite{wmt-2022-machine,wmt-2023,wmt-2024-1}. Despite the current dominance of COMET in system ranking, meta-evaluation studies~\cite{mathur-etal-2020-tangled,freitag-etal-2021-experts,moghe-etal-2025-machine} have highlighted the limitations of relying on a single metric, including problems in reproducing and comparing scores across works \citep{zouhar-etal-2024-pitfalls}, echoing earlier critiques of BLEU~\cite{callison-burch-etal-2006-evaluating,post-2018-call}. Overall, this body of work underscores both the progress achieved in MT evaluation and the persistent challenges in designing metrics that are reliable, robust, and broadly generalizable.

\subsection{Metrics for Speech Translation}
In contrast to MT, evaluation in ST has received comparatively limited attention. While IWSLT has provided the primary benchmark for ST since its first edition in 2004~\cite{akiba-etal-2004-overview}, its focus has largely been on system comparison rather than the development of evaluation metrics, which are substantially inherited as-is from MT. ST, however, introduces two distinctive challenges for automatic evaluation: \textit{i)} segmentation mismatches between speech and text, which complicate direct comparison~\cite{papi-etal-2021-dealing,fukuda22b_interspeech}, and \textit{ii)} the possible absence of gold source transcripts~\cite{cheng21_interspeech,fang-feng-2023-back}, which undermines the applicability of source-aware MT metrics. Although few quality estimation metrics, both source- and reference-free, could be employed (e.g. GEMBA-MQM~\cite{kocmi-federmann-2023-gemba}), they typically rely on large language models and/or API-based services. Such dependence not only entails substantial computational and monetary cost but also raises concerns regarding accessibility, replicability, and long-term stability, as closed-source models can evolve over time and their internal behavior remains opaque. Therefore, they continue to be relatively underused~\cite{larionov2025batchgemba}. Among ST-specific metrics, BLASER~2.0~\cite{dale-costa-jussa-2024-blaser} is, to our knowledge, the only metric capable of source-based evaluation directly from speech. It operates by projecting source audio, reference text, and hypothesis text into a shared multilingual embedding space to assess translation quality. However, \citet{han-etal-2024-speechqe} report that source-aware metrics, such as XCOMET and MetricX, often achieve stronger correlation with human judgments than the BLASER~2.0 end-to-end metric. This suggests that, despite the conceptual advantage of accessing the speech source directly, the current performance of end-to-end speech-based metrics is not yet sufficient to surpass metrics operating on high-quality transcripts. Within IWSLT evaluation campaigns, recent editions have examined the robustness of standard MT metrics for ST under segmentation variations and proposed resegmentation protocols for human evaluation~\cite{sperber-etal-2024-evaluating}. In a similar vein, \cite{post-hoang-2025-effects} analyzed the effect of automatic alignment tools on MT metrics, reporting only minor impacts of segmentation strategies on COMET-based system rankings. However, in contrast to the ongoing efforts in MT, no comparable systematic study or shared task initiative has been conducted in ST, leaving the problem of source-reference misalignments underexplored in the ST domain.

Our work addresses these gaps. Building on advances in MT evaluation, we analyze source-aware metrics in the context of ST, where the speech modality challenges the availability of gold source text and its alignment with reference translations. By systematically assessing their reliability and applicability, we aim to shed light on which evaluation and re-segmentation strategies are most appropriate for ST, in scenarios where the availability of transcripts, or their alignment, cannot be assumed.

\section{Methodology}
\label{sec:methodology}

In this section, we address the two challenges that must be overcome to reliably deploy source-aware metrics in the ST setting: the possible absence of source text (Section~\ref{sec:src-generation}) and the potential segmentation mismatch between the source and the reference translation (Section~\ref{sec:resegmentation}).

\subsection{Synthetic Source Generation}
\label{sec:src-generation}
Our primary goal is to validate the effectiveness of source-aware MT metrics also for ST. Often, however, ST benchmarks do not provide reference transcripts.  To address this limitation, {\it synthetic} source text can be created either by automatically transcribing the input audio ({\bf ASR}) or by back-translating the reference translation into the source language ({\bf BT}). Both methods present distinct advantages and drawbacks, which vary depending on the specific aspect under consideration, as discussed below:

\smallskip

{\bf Coverage and Quality}.
Although there are many ASR models available, even more than MT ones (e.g., on Huggingface\footnote{\url{https://huggingface.co/models}}), their coverage in terms of languages is lower. For example, the only multilingual ASR models supporting more than a few dozen high-resource languages are Whisper~\cite{pmlr-v202-radford23a}, SeamlessM4T~\cite{communication2023seamlessm4tmassivelymultilingual}, XLS-R~\cite{babu22_interspeech}, and OWSM~\cite{owsm}. Their performance, however, is not always excellent across different languages and conditions. On the contrary, there are several multilingual MT models that cover hundreds of languages and thousands of high-performance models specialized in one or a few language pairs. From the first group, we can mention: Madlad-400~\cite{kudugunta2023madlad400}, NLLB~\cite{nllbteam2022languageleftbehindscaling}, mBART-50~\cite{tang2020multilingual}, M2M100~\cite{M2M100}, DeltaLM~\cite{deltalm}, and SeamlessM4T~\cite{communication2023seamlessm4tmassivelymultilingual}. From the second one: the Helsinki-NLP/opus-mt family~\cite{tiedemann2023democratizing}, which altogether covers hundreds of languages and thousands of language pairs,  and IndicTrans2~\cite{gala2023indictrans}, which supports 22 Indic languages. While the comparison between ASR and MT performance must be interpreted with caution, existing studies (see Appendix~A) consistently show that MT systems typically achieve higher overall quality than ASR models.

{\bf Neutrality with respect to third-party system evaluation}.
When evaluating ST systems, using ASR-derived transcripts may introduce bias if the evaluated ST system shares components or training data with the ASR model used to generate the synthetic source. The BT approach avoids this specific dependency, or at least only a residual bias might still occur when the same MT model is used, in opposite directions, for both back-translation and in the evaluated system. This issue is discussed in Section~\ref{sec:exp-src-controlled}.

{\bf Similarity to the missing reference transcript}.
ASR outputs are generally closer to the gold transcripts, particularly in terms of lexicon, syntax, and adherence to the spoken content. BT outputs, instead, may differ more substantially, as they are influenced by the lexical and syntactic choices made in the reference translation, potentially introducing artifacts not present in the original audio. Empirical evidence for this claim comes from multiple results showing ASRs to be more reliable than BTs (see Section~\ref{sec:exp-src-controlled}).

{\bf Alignment with the segmentation of the reference translations}.
For evaluation purposes, the synthetic source needs to be aligned, segment by segment, to the reference translation. The BT ensures alignment by construction. On the contrary, ASR-based synthetic sources have to be (automatically) re-segmented, thus introducing additional processing costs (see next item) and errors that this step typically commits, as discussed in Section~\ref{sec:realistic}.

{\bf Cost}.
The ASR approach requires a speech recognition system and often a re-segmenter, which may introduce significant additional computational costs. The BT approach, instead, requires only a machine translation system, typically less demanding in terms of computational resources. Appendix~B reports on specific experiments performed to compare the two approaches from this perspective. 
\smallskip

In light of the above, the choice between ASR and BT for generating synthetic source text depends on the specific constraints of the evaluation scenario, including computational budget, the availability of effective ASR/BT models, the need for system neutrality, the desired fidelity to the original spoken input, and the ability to effectively perform the automatic re-segmentation, if needed. All these aspects are explored through the experimental results presented in Section~\ref{sec:results} and Appendices B, D, and E, while the final discussion in Section~\ref{sec:discussion} is organized around them.

\subsection{Source Re-Segmentation}
\label{sec:resegmentation}
\label{sec:levenshtein}
\label{sec:refiner}

In realistic settings, automatic translations of input speech are generated from automatically segmented audio. In such a scenario, segment-level evaluation requires aligning the hypothesis against the gold reference text.  To re-segment the translation hypothesis so that its segments match those of the reference translation, the most commonly adopted method is the one proposed by~\namecite{matusov-etal-2005-evaluating}, originally distributed as an executable binary without source code.\footnote{\url{https://www-i6.informatik.rwth-aachen.de/web/Software/mwerSegmenter.tar.gz}} A Python reimplementation, mweralign, was recently made available by~\namecite{post-hoang-2025-effects}, and is the one adopted in this work. For the sake of simplicity, we name it {\bf L-Segmenter}, where the {\bf L} highlights its \textit{intra-lingual} nature: both the generated hypothesis to be re-segmented and the reference text are in the same language (the target one, in this case). The method re-segments the hypothesis so that its Levenshtein distance from the reference is minimized. The optimal segmentation is searched through dynamic programming. Although it is widely adopted by the scientific community, the L-Segmenter algorithm is affected by an intrinsic limitation regarding unaligned boundary words~\cite{post-hoang-2025-effects}. The  upper table of Figure~\ref{fig:resegExample} in Appendix C shows an example of L-Segmenter output, highlighting this problem.

\begin{figure}[t]
\begin{center}
\includegraphics[width=0.99\textwidth]{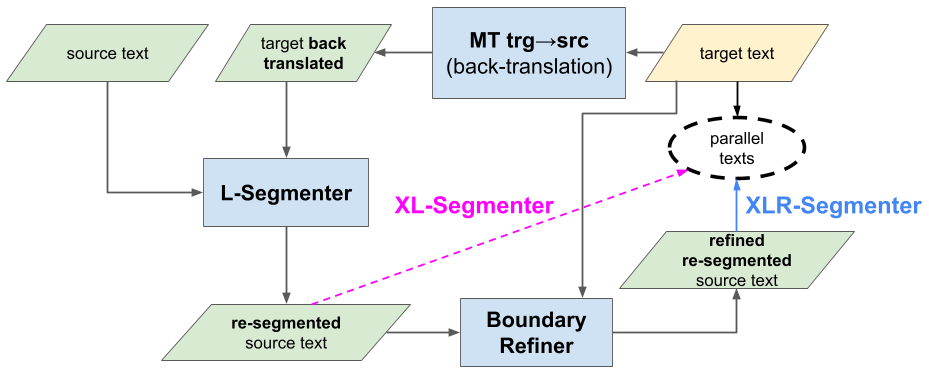}
\end{center}
\caption{Scheme of XL-Segmenter and XLR-Segmenter.}
\label{fig:re-seg}
\end{figure}

The application of source-aware ST metrics extends the segmentation problem to include the source text, which, similar to the translation hypothesis, has to be aligned with the gold reference translation. However, when the source text is not available and needs to be generated with an ASR model, a cross-lingual approach is required to align synthetic transcripts in the source language with target language translations. A possible solution to this problem, out of the reach of the monolingual L-Segmenter, is shown in Figure~\ref{fig:re-seg}. The target text is first translated into the source language, segment by segment, by means of an MT model (back-translation). The source text produced by the ASR model is then aligned by means of L-Segmenter to the back-translation of the target, thus indirectly achieving the cross-lingual alignment. In the figure, we name this segmentation procedure as {\bf XL-Segmenter}, where {\bf X} emphasizes its cross-lingual nature.

\renewcommand{\algorithmicrequire}{\textbf{Input:}}
\renewcommand{\algorithmicensure}{\textbf{Output:}}

\begin{algorithm}[t]
\caption{Pseudo-code of the algorithm for the refinement of segment boundaries}\label{alg:refinement}
\begin{algorithmic}[1]
\Require{${\bf s} = (\ldots, s_i, \ldots ), {\bf t} = (\ldots, t_i, \ldots $)} \Comment{<$s_i,t_i$>: pseudo // sentences}
\Ensure{$\overline{{\bf s}} = (\ldots, \overline{s}_i, \ldots )$}  \Comment{source sentences with refined boundaries}
\Procedure{BoundariesRefinement}{{\bf s}, {\bf t}}\
\For{$i \in [1, ..., len(s) -1]$} \
\State $ \hat{j} \gets {\tt FindOptimalSourceSplit}(s_i + s_{i+1}, t_i + t_{i+1})$ \Comment{+ is the concat operator}
\State $(\overline{s}_i , \overline{s}_{i+1}) \gets {\tt AdjustSource}(\hat{j}, s_i , s_{i+1}) $ \Comment{move words from/to $s_i$ and  $s_{i+1}$ according to $\hat{j}$}
\State ${{\bf s}}_{i+1} \gets \overline{s}_{i+1}$
\EndFor
\State \Return $\overline{{\bf s}}$
\EndProcedure
\end{algorithmic}
\end{algorithm}

However, as mentioned above, L-Segmenter may misplace unaligned words occurring at segment boundaries. Alignment fails because L-Segmenter performs it solely on string matching, ignoring any grammatical considerations (e.g., lexicon, syntax, semantics). Therefore, to refine the boundaries of segments generated by XL-Segmenter, our further extension looks at word alignments built on word embeddings rather than on words' surface form, as the Levenshtein distance does. Embeddings have the advantage of capturing semantic relationships between different words with similar meaning (such as synonyms), thus yielding to more robust and precise alignments.

Algorithm~\ref{alg:refinement} provides a high-level description of the procedure, named {\bf XLR-Segmenter}, where the \textbf{R} indicates the boundary Refinement stage, also shown in Figure~\ref{fig:re-seg}. Given a pair of source/target sentences <$s_i$,$t_i$>, each element is concatenated with its immediate successor ($s_{i+1}$ and $t_{i+1}$, respectively). Then, while keeping the original bipartition of $t_i+t_{i+1}$ fixed, the refinement procedure looks for the optimal splitting point of the source (step~3) and, if it differs from the original one, words are moved accordingly from one segment to the other, updating both $s_i$ and $s_{i+1}$ (steps~4-5).

We implemented two instances of Algorithm 1, which differ in how the optimal splitting point is determined (function {\tt FindOptimalSourceSplit()} at step 3).
The first instance,  {\bf XLR-SimAlign}, uses SimAlign~\cite{jalili-sabet-etal-2020-simalign}\footnote{In our implementation, embeddings are provided by mBERT~\cite{devlin2018bert}, which supports 104 languages.}  to align the words of $s_i+s_{i+1}$ to those of  $t_i+t_{i+1}$. Given this alignment, we count the number of cross-alignments\footnote{A cross-alignment is such if the tokens it connects fall in opposite partitions of the respective strings (the right one of the source and the left one of the target, or vice versa).} for each candidate splitting point $j\in(1, N-1)$, where N is the number of words in $s_i+s_{i+1}$, and select the bipartition that minimises it. Formally, defining $z_i = s_i+s_{i+1}$:
\[
\begin{aligned}
\hat{j}_{\text{SimAlign}} = \argmin_{j\in (1, N-1)} \text{cross-alignments}((z_i[1:j], z_i[j+1:N]), (t_i, t_{i+1}))
\end{aligned} 
\]
The second instance, {\bf XLR-LaBSE}, uses LaBSE~\cite{feng-etal-2022-language} sentence embeddings instead. For each candidate splitting point, we compute the average cosine distance between the resulting source segments and the corresponding target segments, and select the split that minimises it:
\[
\begin{aligned}
\hat{j}_{\text{LaBSE}} = \argmin_{j\in (1, N-1)} \frac{1}{2} ( & \text{cosine\_distance}(\text{LaBSE}(z_i[1:j]), \text{LaBSE}(t_i)) + \\[-10pt]
&    \text{cosine\_distance}(\text{LaBSE}(z_i[j+1:N]), \text{LaBSE}(t_{i+1}))) 
\end{aligned} 
\]
Unlike  {\tt  XLR-SimAlign}, which runs the alignment procedure once per candidate sentence pair,  {\tt XLR-LaBSE} requires computing $2N$ sentence embeddings per input sentence. However, these can be computed in parallel on the GPU, partially offsetting the additional cost (see Appendix~B for a detailed comparison of the computational costs of the two instances).

To showcase the effectiveness of our refinement algorithm, the two tables at the bottom of Figure~\ref{fig:resegExample} in Appendix~C illustrate how the typical boundary issues observed in segments produced by XL-Segmenter are solved by  {\tt XLR-SimAlign}.

\section{Experimental Setting}
\label{sec:exp-setting}

In this section, we describe each building block of our experiments, i.e., data, metrics, systems, and models, providing the motivations supporting each choice.

\subsection{Data}
\label{sec:data}
\begin{table}[t]
\caption{On the {\bf left}: statistics for each section of the MuST-C test set: number of segments, total duration of the English audio, number of (detokenized) source and target words. $(^*)$ For the Chinese side, characters are counted instead of words. On the {\bf right}: statistics for each section of the Europarl-ST test set: total duration (h:mm) of the audio in the source language (top half) and number of (parallel) segments (bottom half)}
\label{tab:data-stat}
\setlength{\tabcolsep}{1.5pt}
 \begin{minipage}{.45\textwidth}
 \begin{tabular}{lcccc}
 \multicolumn{5}{c}{MuST-C}\\
\toprule
trg & \#seg & h:mm & |src| & |trg|\\
\midrule
ar & 2019 & 4:05 & 42.2k & 36.4k \\
cs & 2035 & 3:15 & 36.1k & 29.7k \\
de & 2641 & 4:09 & 46.4k & 44.0k \\
es & 2502 & 4:10 & 46.3k & 42.7k \\
fa & 2113 & 3:37 & 40.0k & 56.7k \\
fr & 2632 & 4:09 & 46.4k & 49.7k \\
it & 2574 & 4:10 & 46.3k & 42.2k \\
nl & 2615 & 4:09 & 46.3k & 42.6k \\
pt & 2502 & 4:10 & 46.2k & 43.2k \\
ro & 2556 & 4:10 & 46.3k & 44.2k \\
ru & 2513 & 4:10 & 46.1k & 36.5k \\
tr & 2408 & 4:10 & 45.9k & 33.1k \\
vi & 2361 & 4:09 & 45.2k & 60.4k \\
zh & 1824 & 4:00 & 39.7k & 74.9k$^*$ \\
\midrule
total & 33295 & 56:31 & 619.1k & 636.3k \\
\bottomrule
\vspace{14.4mm}
\end{tabular}
\end{minipage}
\hspace{-6mm}\begin{minipage}{.55\textwidth}
\begin{tabular}{lccccccccc}
 \multicolumn{10}{c}{Europarl-ST}\\
\toprule
&\multicolumn{9}{c}{trg}\\
\cmidrule{2-10}
src & de & en & es & fr & it & nl & pl & pt & ro \\
\midrule
de &       & 6:03 & 3:16 & 3:10 & 2:57 & 2:58 & 3:08 & 3:11 & 2:48 \\
en & 2:53 &       & 2:55 & 2:47 & 2:43 & 2:51 & 2:51 & 2:53 & 2:31 \\
es & 3:09 & 5:05 &       & 3:05 & 3:08 & 3:05 & 3:01 & 3:05 & 2:36 \\
fr & 2:52 & 4:40 & 2:55 &       & 2:52 & 3:01 & 2:55 & 2:51 & 2:31 \\
it & 2:43 & 5:9 & 2:38 & 2:39 &       & 2:28 & 2:25 & 2:35 & 2:13 \\
nl & 2:30 & 4:01 & 2:24 & 2:21 & 2:14 &       & 2:18 & 2:13 & 2:02 \\
pl & 3:08 & 5:27 & 3:06 & 3:04 & 3:02 & 3:01 &       & 3:06 & 2:29 \\
pt & 3:34 & 6:28 & 3:32 & 3:34 & 3:31 & 3:26 & 3:22 &       & 3:05 \\
ro & 3:37 & 5:38 & 3:33 & 3:24 & 3:32 & 3:33 & 3:25 & 3:32 &       \\
\midrule
de &       & 2631 & 1421 & 1401 & 1217 & 1305 & 1376 & 1387 & 1233 \\
en & 1253 &       & 1267 & 1214 & 1130 & 1235 & 1238 & 1262 & 1095 \\
es & 1114 & 1816 &       & 1082 & 1079 & 1094 & 1059 & 1089 & 910 \\
fr & 1093 & 1804 & 1098 &       & 1046 & 1150 & 1113 & 1100 & 949 \\
it & 922 & 1686 & 885 & 893 &       & 837 & 820 & 871 & 742 \\
nl & 1063 & 1747 & 1014 & 1012 & 890 &       & 967 & 942 & 877 \\
pl & 1284 & 2231 & 1254 & 1259 & 1180 & 1225 &       & 1252 & 993 \\
pt & 1271 & 2286 & 1256 & 1273 & 1205 & 1228 & 1196 &       & 1108 \\
ro & 1231 & 1963 & 1204 & 1157 & 1168 & 1210 & 1164 & 1200 &       \\
\bottomrule
\end{tabular}
\end{minipage}
\end{table}

We experiment with the test sets of two multilingual ST corpora: MuST-C v1.2~\cite{di-gangi-etal-2019-must} and Europarl-ST v1.1~\cite{jairsan2020a}.

\textbf{MuST-C} (Multilingual Speech Translation Corpus) is a large-scale corpus comprising English audio segments aligned with manual transcripts and their corresponding translations into 14 target languages. The corpus is derived from TED Talks,\footnote{\url{www.ted.com}} ensuring coverage of diverse topics and speaking styles, and speaker gender and nationality. Each language-specific section of MuST-C contains hundreds of hours of speech and is segmented at the sentence level, which facilitates supervised training for ASR, MT, and ST models. Our experiments are carried out with all the 14 languages covered by the corpus, namely: Arabic, Czech, Dutch, Farsi, French, German, Italian, Portuguese, Romanian, Russian, Spanish, Turkish, Vietnamese, and Chinese. Table~\ref{tab:data-stat} (left) provides statistics of the MuST-C test set, named tst-COMMON. 

\textbf{Europarl-ST} is a multilingual speech translation corpus built to support research in multilingual and domain-specific settings. It is derived from the Europarl corpus~\cite{koehn-2005-europarl} of European Parliament proceedings, which features textual translations of formal, structured speech across a range of political and legislative topics. The source audio was extracted from publicly available recordings of parliamentary sessions, and the corpus was constructed through careful alignment and filtering to ensure high-quality data for both ASR and ST. Europarl-ST consists of audio recordings in nine European languages (German, English, Spanish, French, Italian, Dutch, Polish, Portuguese, and Romanian), each accompanied by manual transcripts and sentence-aligned translations into the other eight languages, thus covering 72 translation directions. Table~\ref{tab:data-stat} (right) provides statistics of the Europarl-ST test set. 

Since both corpora provide reference transcripts aligned to reference translations, they enable a systematic investigation into how replacing these gold-standard sources with synthetic counterparts, whether ASR-based or BT-based, impacts the reliability of source-aware evaluation metrics. This setup thus allows us not only to assess the potential degradation introduced by synthetic sources but also to anticipate their viability as practical substitutes in real-world scenarios where manual transcripts are unavailable. Moreover, the large total number of segments (33,295 for MuST-C and 88,227 for Europarl-ST), combined with the rich variety of languages, provides a solid empirical basis, ensuring that the results reported in the following sections are grounded on a broad and statistically reliable pool of observations.

\subsection{Metrics}
\label{sec:metrics}

For our study, we selected two metrics that represent the two best families according to the WMT24 Metrics Shared Task~\cite{freitag-etal-2024-llms}. In both the official and Error Span Annotation rankings \citep{kocmi-etal-2024-error}, metrics that incorporate the source text alongside the translation hypothesis and reference consistently achieve the highest performance. We therefore selected COMET-22 (ranked third and widely adopted in the research community) and MetricX-24-Hybrid (ranked first), which are described below:

{\bf COMET}~\cite{rei-etal-2020-comet} is a learned metric based on multilingual pre-trained language models. It operates by encoding the source sentence, MT hypothesis, and human reference using contextualized embeddings (typically from XLM-R~\cite{conneau-etal-2020-unsupervised}), and then scoring the translation quality through a regression model trained on human judgments. Unlike traditional surface-level metrics, COMET captures deeper semantic and syntactic correspondences, yielding stronger correlations with human evaluations across diverse language pairs and domains. Among the several models available, in our experiments we employed the default one, that is {\tt wmt22-comet-da}~\cite{rei-etal-2022-comet},\footnote{\url{https://huggingface.co/Unbabel/wmt22-comet-da}} which covers about 100 languages.

The {\bf MetricX} family consists of state-of-the-art neural models~\cite{juraska-etal-2023-metricx} that leverage large-scale, multilingual transformer architectures to produce quality scores. They are trained on extensive human-labeled data and optimized for robustness across diverse domains, language pairs, and quality ranges. The {\tt Hybrid} variants~\cite{juraska-etal-2024-metricx} combine both reference-based and reference-free evaluation signals, incorporating features from the source, hypothesis, and optionally the reference translation. In our experiments, we employed the {\tt MetricX-24-Hybrid-XL}\footnote{\url{https://huggingface.co/google/metricx-24-hybrid-xl-v2p6}} model, which supports 101 languages.

In some of the experiments discussed below, we also evaluate the quality of ASR and MT, as well as the semantic similarity between source and target texts. For those purposes, we rely on standard metrics, respectively, WER, BLEU, and LASER: 
\smallskip

\noindent
{\bf WER} quantifies the accuracy of ASR systems as the Levenshtein distance, i.e., the minimum number of word-level substitutions, insertions, and deletions needed to transform the system output into the reference transcript, normalized by the total number of words in the reference.  It is computed via jiWER.\footnote{\url{https://github.com/jitsi/jiwer}} Unless otherwise specified, before computation, hypotheses and references are lowercased and punctuation is removed.

\noindent
{\bf BLEU} measures the degree of overlap between a system's output and one or more reference translations by comparing matching $n$-grams. More precisely, it computes the geometric mean of (modified) $n$-gram\footnote{Typically, BLEU considers 1-grams, 2-grams, 3-grams, and 4-grams.} precisions between a translation hypothesis and reference(s), combined with a brevity penalty to account for length differences and discourage overly short hypotheses. It is computed via sacreBLEU \cite{post-2018-call}.\footnote{signature: \texttt{nrefs:1|case:mixed|eff: no|tok:13a|smooth:exp|version:2.0.0}} 

\noindent
{\bf LASER}~\cite{artetxe-schwenk-2019-massively} is not an evaluation metric, but rather a sentence embedding model that produces multilingual vector representations of sentences. We used {\tt laserembeddings 1.1.2},\footnote{\url{https://pypi.org/project/laserembeddings/}} the pip-packaged, production-ready port of LASER, to compute the cosine similarity for any sentence pair, even in different languages. We use it to evaluate the quality of XL-Segmenter and XLR-Segmenter by measuring the similarity between an automatically segmented source text and the manually segmented reference translation. From now on, we refer to the LASER-based similarity score simply as LASER. 

\subsection{ASR and BT Models}
\label{sec:ASR-BT}

For synthetic source generation, we used three ASR models to transcribe the audio source (Whisper, OWSM, and SeamlessM4T), and two MT models for back-translating the reference translations into the source language (MADLAD and NLLB), which are described below:

{\bf Whisper}~\cite{pmlr-v202-radford23a} is a family of open-source models for speech-related tasks trained on 680,000 hours of labeled audio data collected from the web. Built upon a Transformer-based encoder-decoder architecture, Whisper is capable of performing ASR (supporting nearly 100 languages), ST (from those languages into English), timestamp estimation, and language identification. We employed the 1.55B parameters {\tt v3-large multilingual}  model\footnote{\url{https://github.com/openai/whisper}} ({\bf whspr}) to perform the speech transcription.

{\bf OWSM}~\cite{owsm} is a family of multilingual, encoder-decoder speech foundation models, designed for several speech-related tasks, ASR and ST included. For ASR, we used {\tt OWSM v3.1 medium} ({\bf owsm})~\cite{owsm-v31},\footnote{\url{https://huggingface.co/espnet/owsm\_v3.1\_ebf}}  a model with 1.02B parameters in total, trained on 180k hours of public speech data and supporting the transcription for 151 languages.

{\bf SeamlessM4T}~\cite{communication2023seamlessm4tmassivelymultilingual} is a foundation all-in-one Massively Multilingual and Multimodal Machine Translation model supporting several speech-related tasks, ASR and ST included, in nearly 100 languages. For ASR, we used {\tt SeamlessM4T v2 large} ({\bf smlss}),\footnote{\url{https://huggingface.co/facebook/seamless-m4t-v2-large}} a transformer model with 2.3B parameters.

{\bf MADLAD} is a family of large-scale, pre-trained models for MT. These models are designed to support translation across over 400 languages, many of which are low-resource and underrepresented in existing MT systems. The training corpus for MADLAD models comprises approximately 18 billion sentence pairs, mined and filtered from a wide variety of multilingual web and public sources to ensure broad linguistic and domain diversity. In our experiments, we used \texttt{MADLAD-400-3B-MT} ({\bf mdld})~\cite{kudugunta2023madlad400}, a 3B-parameters MT model capable of performing bidirectional translation between all supported languages. The model is based on a Transformer architecture and is trained with a multilingual objective to enable zero-shot and few-shot generalization across diverse language pairs. 

{\bf NLLB}~\cite{nllbteam2022languageleftbehindscaling} is a family of multilingual MT models, aimed at enabling translation across a wide spectrum of languages, including low-resource and underrepresented ones. \texttt{NLLB-200-3.3B} ({\bf nllb}) is a 3.3-billion-parameter Transformer-based model that supports direct translation between any pair of the 200 covered languages. 

\smallskip

The average quality of the three ASR models and of the two MT models for the BT task is provided in Table~\ref{tab:BT-results}.

\begin{table}[ht]
\caption{ASR ({\bf left}) and MT (for the BT task, {\bf right}) performance on the MuST-C and Europarl-ST test sets}
\label{tab:BT-results}
\setlength{\tabcolsep}{3pt}
\begin{minipage}{.31\textwidth}

\begin{tabular}{l|c|c}
\toprule
ASR & \small MuST-C & \small EP-ST \\
 model      & \small\%WER$\downarrow$ & \small \%WER$\downarrow$ \\
\midrule

{\tt whspr} & 6.97 & 10.40 \\
{\tt owsm} & 9.37 & 20.71 \\
{\tt smlss} & 18.53 & 9.84\\
\bottomrule
\end{tabular}

\end{minipage}
\begin{minipage}{.70\textwidth}
\begin{tabular}{l|ccc|ccc}
\toprule
MT       & \multicolumn{3}{c|}{MuST-C} & \multicolumn{3}{c}{Europarl-ST} \\
model & \small BLEU & \small COMET & \small MetricX & \small BLEU & \small COMET & \small MetricX  \\
for BT       &\small(0-100$\uparrow$)&\small(0-1$\uparrow$)&\small(0-25$\downarrow$)&\small(0-100$\uparrow$)&\small(0-1$\uparrow$)&\small(0-25$\downarrow$)\\
\midrule
{\tt mdld}  & 38.74 & 0.8562 & 3.398 & 31.40 & 0.8810 & 2.439 \\
{\tt nllb}  & 39.94 & 0.8510 & 3.326 & 25.63 & 0.8627 & 2.891 \\
\bottomrule
\end{tabular}
\end{minipage}
\end{table}

Starting from ASR, although performance is overall excellent, we observe cases where results are particularly low, specifically those of SeamlessM4T on MuST-C and of OWSM on Europarl-ST.  In the first case, SeamlessM4T's ASR results on MuST-C tend to be systematically less accurate compared to Whisper and OWSM.\footnote{While this result is surprising, it is beyond the scope of this work to identify the causes of this behavior, which we leave to future works.} Concerning OWSM, it is quite strong in transcribing English (WER$_{\tt os}$=12.70\% vs. WER$_{\tt wh}$=11.33 and WER$_{\tt sm}$=11.24) but rather weak in other languages, particularly Portuguese (WER=39.35\%), Romanian (WER=27.69\%) and Polish (WER=27.35\%). This explains its high average WER on Europarl-ST. Regarding BT, the values of COMET and MetricX are all very good, ensuring an extremely high quality of the synthetic sources generated by MADLAD and NLLB. However, while on MuST-C the two models are substantially equivalent, MADLAD turns out to be significantly better than NLLB on Europarl-ST.

\subsection{ST Models}
\label{subsec:st-models}
To ensure the broad validity of our findings, we complemented the variety of data and domains (Section~\ref{sec:data}) with a heterogeneous set of ST systems, differing in both architecture and performance. We opted for three direct systems among those capable of covering most of the translation directions present in our two benchmarks, and three cascaded systems, where the ASR instance of the direct ST models was coupled with MADLAD, the best performing state-of-the-art MT model among the two involved in our investigation.\footnote{Direct ST systems map input speech directly into the target language text in a single end-to-end model, without generating intermediate transcripts. Cascade ST systems operate in two stages: first, transcribing the source speech into text with an ASR model and then translating that text into the target language with an MT model. See~\cite{bentivogli-etal-2021-cascade} for a comparison of the two architectures.} Specifically, the six ST systems are:

\begin{itemize}
\item {\bf whisperST}:  We employed the same Whisper model used to perform ASR to also generate the direct translation from non-English speech into English text.

\item {\bf whspr+mdld}: Cascade of the Whisper model, used as ASR for transcribing the speech of the original audio recording, and MADLAD for translating the transcripts into the target language. 

\item {\bf owsmST}: 
We employed the same OWSM model used to perform ASR to also generate the direct translations.

\item {\bf owsm+mdld}: Cascade of the OWSM model for ASR and MADLAD for MT.

\item {\bf seamlessST}: We employed the same SeamlessM4T model used to perform ASR to also generate the direct translations.

\item {\bf smlss+mdld}: Cascade of the SeamlessM4T model for ASR and MADLAD for MT.
\end{itemize}

\begin{table}[ht]
\caption{Overview of data used: number of language pairs and of segments each system worked on}

\label{tab:system-data}
\centering
\begin{tabular}{l|cc|cc}
\toprule
\multirow{2}{*}{system} & \multicolumn{2}{c|}{MuST-C} & \multicolumn{2}{c}{EuroparlST}  \\
& |src-tgt| & \#seg & |src-tgt| & \#seg \\
\midrule
 whisperST  & -  & -      & 8  & 16,164 \\ 
 whspr+mdld  & 14 & 33,295 & 72 & 88,227 \\
 owsmST     & 14 & 33,295 & 64 & 79,294 \\
 owsm+mdld  & 14 & 33,295 & 72 & 88,227 \\
 seamlessST & 14 & 33,295 & 72 & 88,227 \\
 smlss+mdld  & 14 & 33,295 & 72 & 88,227 \\
 \midrule
 total      & 70 & 166,475 & 360 & 448,366 \\
\bottomrule
\end{tabular}
\end{table}

Table~\ref{tab:system-data} shows the main statistics regarding the sections of the benchmarks actually covered by each of the six ST systems. Since {\tt whisperST} supports ST only into English, and MuST-C includes only  English speech, the system cannot be used in experiments on that corpus. Europarl-ST, instead, can be employed only for the eight to-English directions. {\tt owsmST} does not cover the translation into Polish; therefore, the eight to-Polish tasks are excluded from experiments.

\begin{table}[ht]
\caption{ST performance of each system on the MuST-C and the Europarl-ST test sets. For the cascade architectures,  the quality of the automatic transcription is also provided, in terms of WER. For each metric, values are arithmetically averaged on all  language pairs covered by each system (see Table~\ref{tab:system-data})}
\label{tab:ST-results}
\setlength{\tabcolsep}{3pt}
\begin{tabular}{l|cccc|cccc}
\toprule
       & \multicolumn{4}{c|}{MuST-C} & \multicolumn{4}{c}{Europarl-ST} \\
system & WER & BLEU & COMET & MetricX & WER & BLEU & COMET & MetricX  \\
       &\%$\downarrow$&(0-100$\uparrow$)&(0-1$\uparrow$)&(0-25$\downarrow$)&\%$\downarrow$&(0-100$\uparrow$)&(0-1$\uparrow$)&(0-25$\downarrow$)\\
\midrule
whisperST  &   -   &   -   &    -   &   -   &       & 28.89 & 0.8091 & 5.636  \\
whspr+mdld  & 6.97  & 25.79 & 0.8167 & 3.963 & 10.40 & 29.79 & 0.8524 & 3.608  \\
owsmST     &       & 16.18 & 0.6504 &10.798 &       & 3.87  & 0.5292 & 15.005 \\
owsm+mdld  & 9.37  & 25.49 & 0.8041 & 4.543 & 20.71 & 26.23 & 0.8028 & 5.583  \\
seamlessST &       & 22.34 & 0.7833 & 5.103 &       & 22.04 & 0.7977 & 5.449  \\
smlss+mdld  & 18.53 & 24.13 & 0.7974 & 4.553 & 9.84  & 30.26 & 0.8542 & 3.469  \\
\bottomrule
\end{tabular}
\end{table}

Table~\ref{tab:ST-results} provides the overall performance of the six ST systems on the two benchmarks. To better contextualize these values,we compare the MuST-C scores with those reported by \namecite{tsiamas-etal-2024-pushing}, which spans 35 recent systems. The best BLEU, averaged on 8~language pairs, is 33.2. Our {\tt whspr+mdld} system  scores 31.8 on the same set of language pairs, and it would rank fifth. In general, cascaded systems perform better than direct ones. The only competitive direct system is {\tt whisperST}, while {\tt owsmST}'s average performance is particularly low, especially on Europarl-ST. In fact, while {\tt owsmST} translates the English speech section of Europarl-ST similarly to how it does in MuST-C (the average BLEU on the seven Europarl-ST from-English pairs is 15.50, compared to the average of 16.18 on MuST-C), it performs poorly on all other language pairs, especially those not involving English.

All in all, our choice of ST models yields broad diversification not only in terms of architecture but also in terms of performance, including cases of particularly low quality of transcripts and translations.

\subsection{Metric Evaluation Criterion}

As our goal is to verify the effectiveness of using synthetic sources to compute source-aware metrics, we need to evaluate whether the scores obtained through synthetic sources are reliable or not. To this aim, we consider the source-aware metrics with manual transcripts as our gold standard, relying on their demonstrated high correlation with human judgment, and we compute the Pearson correlation~\citep{pearson1895note} of the scores obtained with synthetic sources with those obtained with manual transcripts. We choose Pearson correlation over alternative methods such as Kendall and Spearman because we care not only about the returned ranking but also about the magnitude of the differences in scores.

The correlation scores are computed independently for the two benchmarks (see Section \ref{sec:data}), for each metric (see Section \ref{sec:metrics}), for each method used to generate synthetic sources (see Section \ref{sec:ASR-BT}) and for each ST system (see Section~\ref{subsec:st-models}). For instance, if we use the Whisper ASR to generate the synthetic source, MuST-C as a benchmark, COMET as a metric, and SeamlessM4T as an ST model, we generate the outputs with SeamlessM4T for all the 33,295 segments of MuST-C, then we compute the segment-level COMET scores both using the manual (gold) transcript and the synthetic one generated by Whisper, and finally compute the correlation between the pairs of segment-level scores. The resulting correlation score tells us whether the synthetic source reflects the behavior obtained with the manual source (if the score is close to 1) or it changes to be unrelated to the ``correct'' score (if the score is close to 0). Following~\cite{cohen1988spa}, we consider correlations $>0.80$ as very strong. We do not report in the main body of this work the correlations between system rankings and scores on the whole test sets, as we noticed that these correlations are always very high ($>0.99$) for all methods and metrics, as we show in Appendix~D. 

From here on, for simplicity, we name {\it standard} the metric that uses the manual transcript, and either {\it BT}, {\it ASR} or simply {\it  synthetic} the one with, respectively, the BT source, the ASR source or any of them.

\section{Experiments}
\label{sec:results}
In this section, we first investigate the impact of using synthetic sources in the simplified scenario in which the alignment between the reference translation and the source audio is provided (Section~\ref{sec:exp-src-controlled}). This preliminary investigation has a two-fold objective: to assess whether synthetic sources compromise the reliability of source-aware metrics, and to compare the ASR and BT options under optimal conditions for ASR (without possible realignment errors). We complement this investigation with an analysis of the factors driving the superiority of one source generation method over the other (Section~\ref{sec:synth-source-analysis}). Then, we move to the scenario in which the reference translation is not aligned to the source audio. In Section~\ref{sec:reseg-controlled}, we evaluate the effectiveness of our re-segmentation method on the manual transcripts, thus preventing errors made by ASR systems from influencing the results of the synthetic source text. Lastly, we compare the BT and ASR approaches in a realistic scenario, where the ASR source is realigned to the reference translation using our proposed method (Section~\ref{sec:realistic}). The two following sections extend the main evaluation in two complementary directions: Section~\ref{sec:additional-exps_low-resource} explores a low-resource setting involving the Bemba-English language pair, while Section~\ref{sec:additional-exps_DA} provides a direct validation of the synthetic metrics against human quality judgments.

\subsection{Comparison of Synthetic Sources with Known Audio-Reference Alignments}
\label{sec:exp-src-controlled}

\paragraph{\bf  Outline of experiments} In this first set of experiments, we exploit the {\it manual audio segmentation}, which ensures the alignment of the ASR transcripts with the reference translations. This controlled setup allows us to perform accurate comparisons that avoid the influence of the automatic re-segmentation of the transcripts. 

\begin{table}[t!]
\caption{Pearson correlations between the COMET/MetricX scores of each system on MuST-C, computed (i) by using as the source the reference transcript (i.e., in the standard way), and (ii) by using synthetic sources (either automatic transcripts or back-translations of the reference target sentences). Each reported correlation coefficient is the average of the correlation coefficients computed separately for each language pair. \textdagger: correlations potentially biased, as the synthetic COMET/MetricX variant relies on ASR transcripts produced by a system related to the evaluated system (for a discussion, see below the key observation {\it Biased conditions}). The heat map colouring is applied independently for each metric (COMET and MetricX) and covers only the synthetic source rows, excluding {\tt shuff}, {\tt COMET-ref}, and {\tt MetricX-ref}. Within each metric, the heat map over individual systems (yellow for COMET, green for MetricX) is separate from that over their average (grey)}

\label{tab:corr-mustc}
\centering
\includegraphics[width=1.0\textwidth]{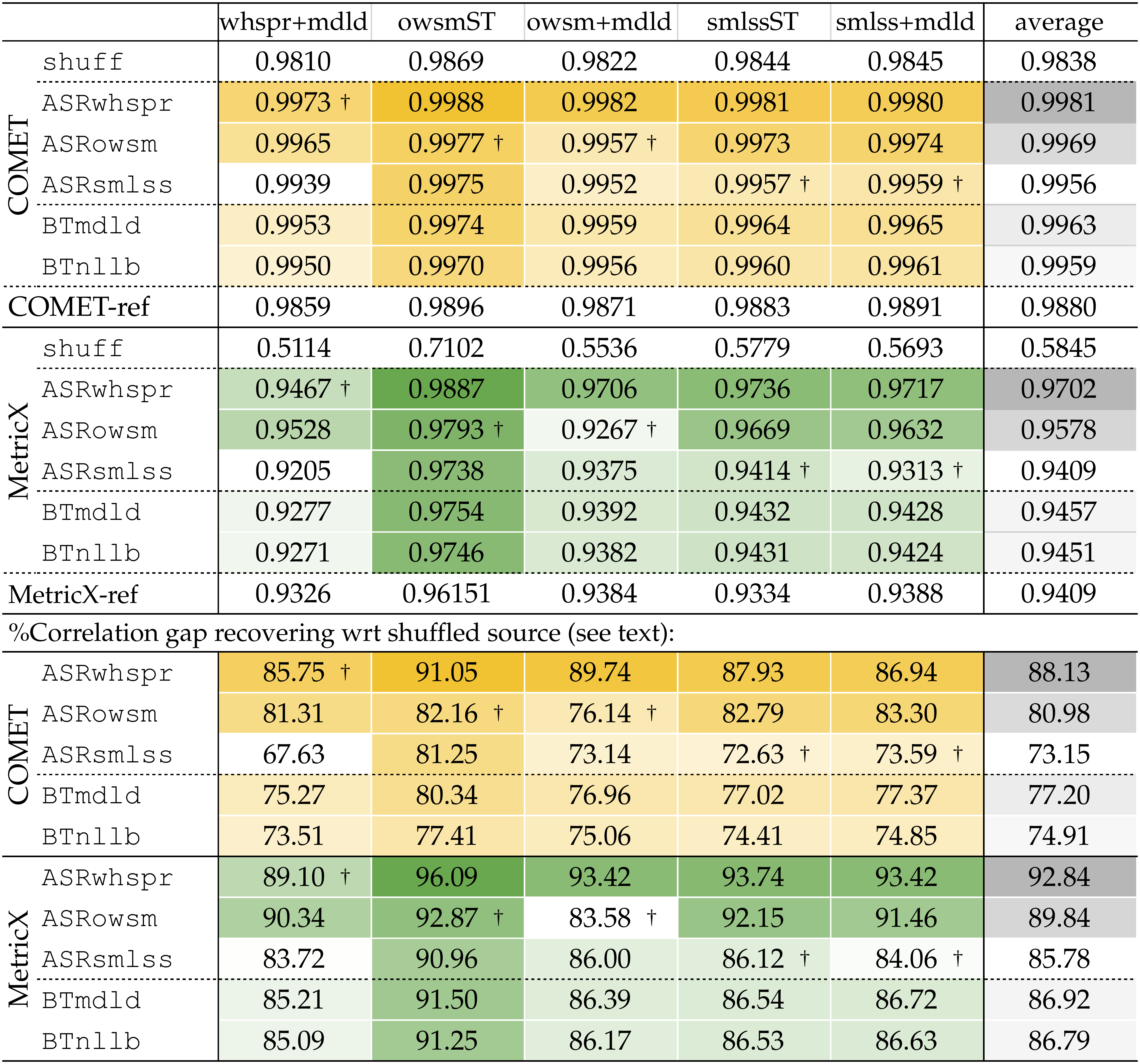}
\end{table}

\paragraph{\bf Results} The upper part of Tables~\ref{tab:corr-mustc} and~\ref{tab:corr-epst} provides, for both corpora (MuST-C and Europarl-ST) and both source-aware metrics (COMET and MetricX), the correlations scores
of the five ST systems capable of translating from English (see Table~\ref{tab:system-data}). 
To define a lower bound for the correlation values, we include the correlations with the scores obtained by randomly shuffling the segments of reference transcripts ({\tt shuff}), that is, by completely inhibiting any relationships between the source and the two translations (hypothesis and reference). As an additional validation of the usefulness of synthetic sources, we also report the correlations obtained with the variants of the two metrics that do not use the source, namely COMET-ref and MetricX-ref. The lower part of the tables indicates the extent to which the correlation gap between random and manual sources is recovered by exploiting synthetic sources. For each of the five ST systems and each metric, either COMET or MetricX, it is computed as follows:
$$
{\tt \%Correlation \ Gap \ Recovering } = 100 \times \frac{r({\tt <syntetic \ source>})-r({\tt shuff})}{1-r({\tt shuff})}
$$
\noindent where {\tt <syntetic source>} is one of the three ASR and two BT synthetic sources. For example, for {\tt smlssST} the correlation between MetricX with shuffled source and the standard MetricX is 0.5779, that of MetricX with {\tt BTnllb} source is 0.9431, then the \%{\tt Correlation Gap Recovering} is $100\times(0.9431-0.5779)/(1-0.5779) = 86.53$. 
In other words, by setting the lower bound of the correlation obtained with the shuffled source to 0, and the upper bound of the correlation obtained with the manual source to 100, the correlation value obtained with the synthetic sources is positioned proportionally.

\smallskip

\noindent
{\bf Key observations and takeaways.} 
\smallskip

\noindent
{\it -- Both synthetic sources are effective.} Overall, all correlation values are very high. On MuST-C, the correlations for COMET are always above 0.99, while for MetricX, there is only one case where the correlation is slightly below 0.93. On Europarl-ST, the minimum values are 0.9784 for COMET and 0.8882 for MetricX. 
In any case, values above 0.80 indicate a very strong correlation \cite{cohen1988spa} between the standard and synthetic metrics, meaning that any proposed synthetic source is an effective substitute for the manual transcript. Finally, it is worth noticing that on average, the correlations of COMET-ref and MetricX-ref are lower than those obtained with any synthetic metric.

\smallskip
\noindent
{\it -- ASR is better than BT.} In general, with ASR sources, the correlations are higher than with BT, with few exceptions when the sources are generated by the worst ASRs (from Table~\ref{tab:ST-results}, SeamlessM4T for MuST-C and OWSM for Europarl-ST). Therefore, ASR sources appear to be more effective than BT ones as long as their quality is good enough, an aspect we explore in Section~\ref{sec:synth-source-analysis}. 

\begin{table}[ht!]
\caption{Pearson correlations between the COMET/MetricX scores of each system on Europarl-ST, computed (i) by using the reference transcript as the source (i.e. in the standard way), and (ii) by using synthetic sources (either automatic transcripts or back-translations of the reference target sentences). Each reported correlation coefficient is the average of the correlation coefficients computed separately for each language pair. For \textdagger~(correlations potentially biased) and heat map colouring see caption of Table~\ref{tab:corr-mustc}}

\label{tab:corr-epst}
\setlength{\tabcolsep}{2pt}
\includegraphics[width=1.0\textwidth]{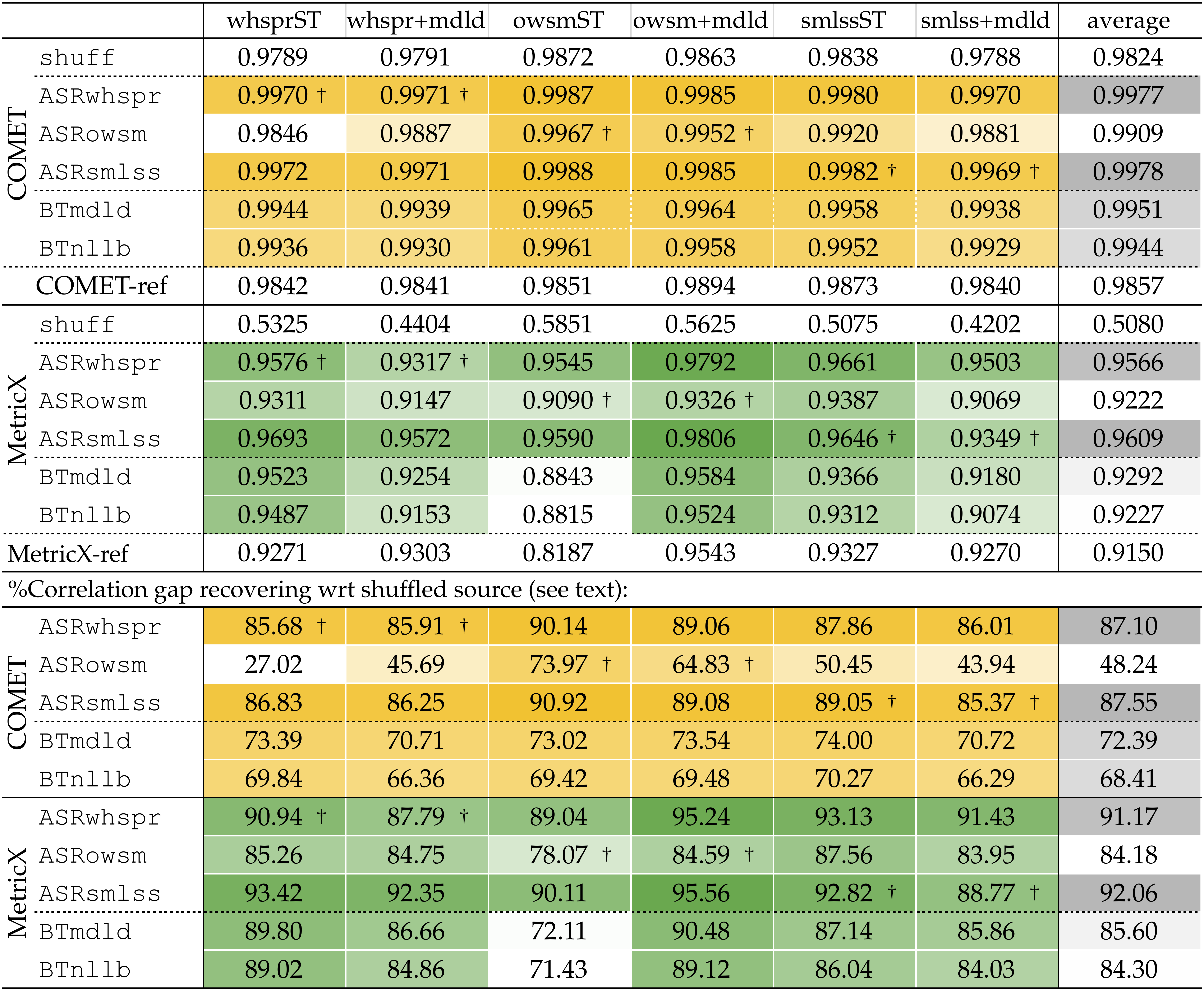}
\end{table}

\smallskip
\noindent
{\it -- The correlation gap is almost fully recovered.} Looking at the percentage recovery instead of absolute correlations, values are always very high for MetricX (ranging from a minimum of 67.63 to a maximum of 97.07), and high for COMET, with the exception of {\tt ASRowsm} on Europarl-ST, due to the poor performance of the recognizer. These numbers confirm the general effectiveness of synthetic sources as a substitute of manual sources, although conditioned to transcription quality. 

\smallskip
\noindent
{\it -- Random sources have different impact.} COMET's correlations with the randomly shuffled source are surprisingly high (the lowest observed value is 0.9794). We hypothesize that this metric was trained to use the source but gives it a definitely smaller weight than the reference translation in evaluating the quality of the translation hypothesis. For MetricX, the source contribution to the final score is significantly higher, as using a source uncorrelated with the targets causes the correlations to plunge even to 0.4185. This suggests that COMET may attenuate any differences observed on MetricX in experiments involving the source text, which should be taken into account when interpreting the results.

\smallskip
\noindent
{\it -- Biased conditions.} The final observation concerns specific cases where the ASR source is linked to the ST system under evaluation, that is, when the ASR  source is produced by a system directly involved in the generation of the ST hypotheses, such as the ASR component of an ST cascade (e.g., {\tt whspr+mdld}) or the ST system itself (e.g., {\tt owsmST}). In these biased conditions (see the scores marked with a dagger), we observe correlation values that tend to be lower. Our hypothesis is that, in these cases, the metric scores are artificially inflated, regardless of whether the transcript and translation are actually correct. In particular, for cascaded systems, since the output translation is generated using the same transcript as input of the MT component, the metric with ASR source actually assesses the MT component rather than evaluating the whole pipeline. The effect is particularly evident for MetricX, which, as already noted, relies more heavily on the source compared to COMET. In light of this observation, biased cases will be excluded from the aggregate results we present from now on.

\smallskip

This first set of experiments indicates that the proposed synthetic replacements of the original sources are extremely effective, yielding synthetic versions of the two metrics that correlate very strongly with their standard versions. The observation that COMET consistently correlates well, even by replacing the original source with a randomly shuffled version of it, leads us to continue the investigation only on MetricX, which, by contrast, proved to be more sensitive to the contribution of the source.

\subsection{Insights into Synthetic Source Comparison}
\label{sec:synth-source-analysis}

\paragraph{\bf  Outline of experiments} In this set of experiments, we examine whether the previously observed superiority of ASR-based sources over the BT-based ones holds systematically or only under specific conditions. Specifically, we investigate how the effectiveness of synthetic sources is influenced by three key factors:
(i) their {\it quality}; (ii) the {\it languages} involved in the evaluation; and (iii) the {\it architecture} of the ST system being evaluated.

\begin{figure}[t!]
\begin{center}
\includegraphics[width=0.5\textwidth]{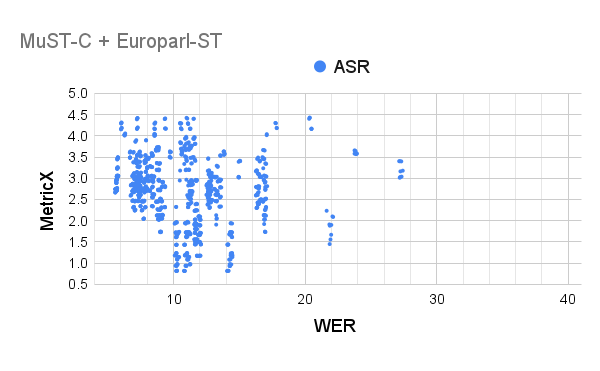}\includegraphics[width=0.5\textwidth]{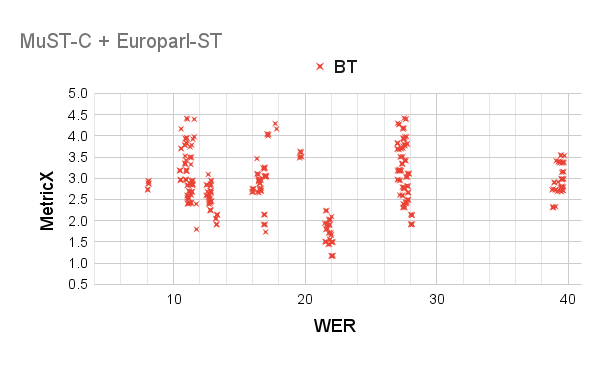}
\end{center}
\caption{The plane of these scatter charts is defined by the WER and MetricX scores of the ASR and BT sources, respectively.  For all possible comparisons between the MetricX correlation with the ASR source and that with the BT source, computed on all language pairs of the two corpora and for all ST systems, the two charts show where the cases in which it was preferable to use the ASR (on the left) or the BT (on the right) as a source for the computation of MetricX are placed in that plane. Biased ASR MetricXs, i.e., those of ST systems that are somehow involved in the generation of the ASR source of the metric, are excluded. The total number of points is 1672, 1315  on the left (ASR wins, 78.6\%), 357 on the right (BT wins, 21.4\%). A random 1\% change was applied to all values to avoid the overlapping of points and make all of them visible.} 
\label{fig:plotUnion}
\end{figure}

\begin{figure}
[ht]
\begin{center}
\includegraphics[width=0.5\textwidth]{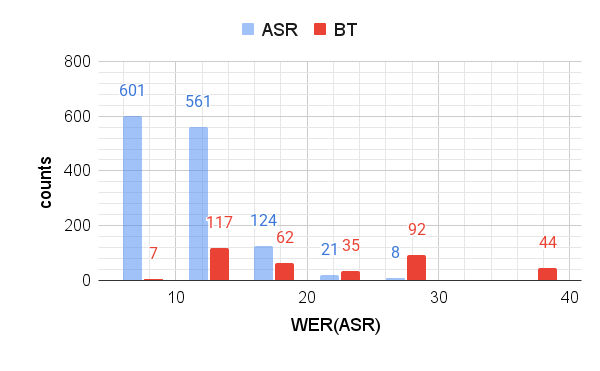}\includegraphics[width=0.5\textwidth]{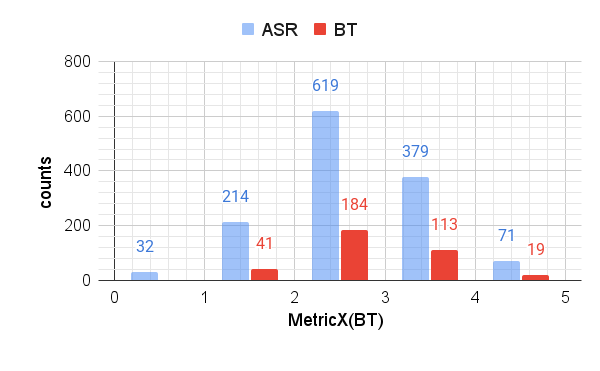}
\end{center}
\caption{For all possible comparisons between the MetricX correlation with the ASR source and that with the BT source, computed on all language pairs of the two corpora and for all ST systems, the histograms illustrate the distribution of cases in which the standard MetricX shows a higher correlation with MetricX using either the ASR or the BT as source input. Biased ASR MetricXs, i.e., those of ST systems that are somehow involved in the generation of the ASR source of the metric, are excluded. The left chart reflects this distribution as a function of transcription quality (WER), while the right chart does so with respect to (back-)translation quality (MetricX). }
\label{fig:histoUnion}
\end{figure}

\paragraph{\bf Results} Concerning the possible dependency on the {\it quality}, Figure~\ref{fig:plotUnion} plots the quality of the ASR source (measured by WER) against the quality of the BT (measured by MetricX): for each possible ASR vs. BT pair, if the winner (higher correlation with the standard metric) is ASR, the (blue) point is placed on the left chart; if it is BT, the (red) cross mark is placed on the right chart. Figure~\ref{fig:histoUnion} shows the same distributions from another perspective, where the positioning and the height of the histogram bars (proportional to the counts) allow for a more immediate visualization of the dependence (if any) of the effectiveness of the ASR and BT sources on their quality. It is evident that the effectiveness of ASR sources strictly depends on their quality (ASR is the winner only if the WER is low), while there is no such regularity for BT. Since both figures suggest the WER=20\% threshold as particularly discriminatory, Table~\ref{tab:winners-union-controlled} collects the counts broken down by that value. It turns out that ASR is preferable in 78.6\% of the cases, but the percentage increases up to 87.4\% when WER$\le$20\%. Above that threshold, BT wins 85.5\% of times. In other words, we can state that in the presence of an ASR with WER not exceeding 20\%, the probability that it is a better substitute of the ideal source than any BT is over 87\%, otherwise the choice of a BT is winning at almost 86\%. From the {\it language} perspective, Table~\ref{tab:winsPerLang} shows a breakdown of the total counts of Table~\ref{tab:winners-union-controlled} by language pair. In most cases, ASR wins with the only exception of Polish as the source language, where BT  slightly prevails over ASR. Overall, the results show no dependence on the specific language pair, confirming the previous finding that ASR sources are generally better substitutes of manual sources regardless of the languages involved in the translation process. In relation to the ST system {\it architecture}, Table~\ref{tab:winsPerSys} shows a breakdown of the total counts of Table~\ref{tab:winners-union-controlled} by system. First, ASR sources are consistently the best option across all the systems. Comparing the two types of ST architectures (cascaded and direct), their prevalence is slightly more marked in direct (575 vs. 97, i.e., 86\% vs. 14\%) than in cascaded systems (740 vs. 260, i.e., 74\% vs. 26\%). Despite this, the greater effectiveness of ASR sources compared to BT sources remains confirmed regardless of the architecture of the ST system being evaluated.

\newpage
\noindent
{\bf Key observations and takeaways.}

\smallskip
\noindent
{\it -- The quality of ASR impacts the effectiveness of synthetic metrics.} Our results confirm that, in general, ASR sources work better than BT sources if the WER remains below 20\%, otherwise BT is the best choice.

\smallskip
\noindent
{\it -- Our findings are consistent across languages.} There is no evidence that the specific languages involved in the speech translation process need to be taken into account when choosing between ASR or BT sources.

\smallskip
\noindent
{\it -- Systems' architecture does not impact.} Similarly, the architecture of the ST system being evaluated does not appear to impact the choice either.

\smallskip
These experiments show that ASR sources are preferable to BT sources, provided their WER is sufficiently good ($\le$20\%). They also highlight that the choice between ASR and BT is not affected by the languages involved in the ST process nor by the architecture of the ST systems being evaluated.

\begin{table}[ht]
\caption{For all possible comparisons between the MetricX correlation with the ASR source and that with the BT source, computed on all language pairs of the two corpora and for all ST systems, the total number of wins per synthetic source type (ASR -- whisper, owsm, seamless -- or BT - madlad, nllb) is given here. Biased ASR MetricXs, i.e. those of ST systems that are somehow involved in the generation of the ASR source of the metric, are excluded}
\label{tab:winners-union-controlled}
\setlength{\tabcolsep}{1.7pt}
\begin{tabular}{l|cccc|cccc|cccc}
\hline
\multirow{3}{*}{corpus} 
& \multicolumn{4}{c|}{WER$_{\tt ASR}\le$20\%} & \multicolumn{4}{c|}{WER$_{\tt ASR}>$20\%} & \multicolumn{4}{c}{total} \\
& \multicolumn{2}{c}{ASR wins} & \multicolumn{2}{c|}{BT wins}& \multicolumn{2}{c}{ASR wins} & \multicolumn{2}{c|}{BT wins}& \multicolumn{2}{c}{ASR wins} & \multicolumn{2}{c}{BT wins} \\
& counts & \% & counts & \% & counts & \% & counts & \% & counts & \% & counts & \% \\
\hline
MuST-C       & 206  & 78.6 & 56  & 21.4 & 18  & 100.0 &  0  & 0.0  & 224  & 80.0 & 56  & 20.0 \\
Europarl-ST  & 1080 & 89.3 & 130 & 10.7 & 11  & 6.0   & 171 & 94.0 & 1091 & 78.4 & 301 & 21.6 \\
\hline
total        & 1286 & 87.4 & 186 & 12.6 & 29  & 14.5  & 171 & 85.5& 1315 & 78.6 & 357 & 21.4 \\
\hline
\end{tabular}
\end{table}

\begin{table}[ht]
\caption{ASR/BT win counts per language pair, visualised as a heat map of proportional pie charts. Each cell (target language × source language) displays a bicolour ellipse whose area scales with the total win count for that pair; the blue sector represents ASR wins and the red sector BT wins, with their respective counts printed inside the ellipse when space permits, and outside otherwise. Marginal stacked bars on the right and at the bottom aggregate wins per target language and per source language, respectively, with counts annotated inline or adjacent to each segment. The pie chart in the bottom-right corner summarises grand totals}
\label{tab:winsPerLang}
\centering
\includegraphics[width=1.0\textwidth]{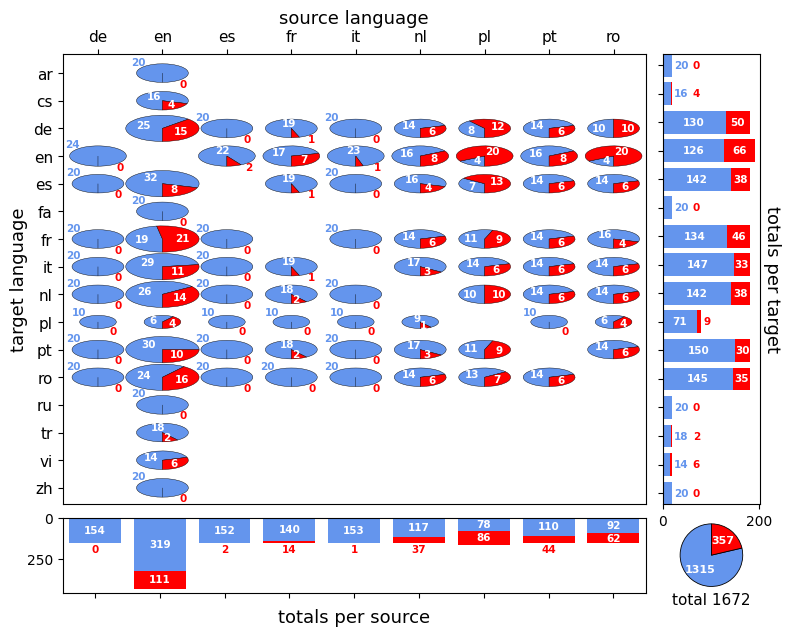}
\end{table}

\begin{table}[ht]
\caption{ASR/BT win counts per speech translation system and decoding architecture, visualised as a heat map of proportional pie charts. Each cell (system × architecture) displays a bicolour ellipse whose area scales with the total win count; blue denotes ASR wins, red denotes BT wins, with counts annotated inside or outside the ellipse according to available space. Marginal stacked bars on the right aggregate wins per architecture. The pie chart on the right summarises grand totals}
\label{tab:winsPerSys}
\centering
\includegraphics[width=1.0\textwidth]{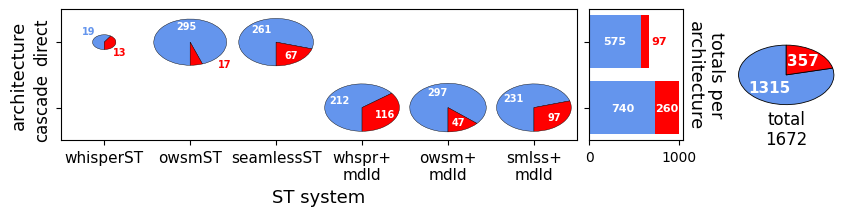}
\end{table}

\subsection{Source Re-segmentation of Manual Transcripts}
\label{sec:reseg-controlled}

\paragraph{\bf Outline of experiments}

In the previous experiments, we assumed that the alignment between the reference translation and the source audio is known. As discussed in Section~\ref{sec:intro} (RQ3), this represents a simplistic assumption that might not hold in real-world conditions, but allows performing the comparisons, avoiding the influence of the automatic re-segmentation of the transcripts. Similarly, to evaluate the effectiveness of our approach for re-segmenting the synthetic sources (see Section~\ref{sec:resegmentation}), in this section we assess its ability to re-segment the {\it manual transcripts}. This controlled setup enables us to precisely measure how well our re-segmenters (Figure~\ref{fig:re-seg}) can recover the original segmentation, preventing errors made by ASR systems from influencing the results. Specifically, we first artificially modify the gold, reference-aligned segmentation of the manual transcripts by randomly splitting the transcripts into segments containing 5 to 100 detokenized words. Then, we apply XL-Segmenter, XLR-SimAlign and XLR-LaBSE to the randomly segmented sources\footnote{Since our re-segmentation procedure requires generating BT of the original translations, we perform this step using both MT models considered in our study, MADLAD and NLLB (Section~\ref{sec:ASR-BT}).} and evaluate how their outputs deviate with respect to the original manual segmentation of the dataset. To this aim, we compute LASER scores between the source text in the new segments and the corresponding segments of the reference translations.\footnote{We compare our outputs with the references instead of the original source text, as the benchmarks used in our experiments contain some misalignments between the source and target text \cite{ouyang-etal-2022-impact}, most likely because
these alignments are automatically produced. While \namecite{ouyang-etal-2022-impact} showed that these misalignments do not compromise the reliability of the evaluation of ST quality, they might have an impact when it comes to assessing the correct alignment between texts. To avoid this affecting our evaluation, we assess the alignment of the source text against the reference target text.} Since LASER scores are based on the cosine similarity between the two sentence embeddings, the closer these scores are to 1, the better the automatic re-segmentation works.

\paragraph{\bf Results} 
Table~\ref{tab:in-vitroReseg} shows the LASER scores computed between the manually segmented reference translations and differently segmented reference transcripts. The LASER scores for the manual segmentation of the two benchmarks evaluate the  correspondence between the reference translations and the manual transcripts, without these scores being affected by segmentation differences. They therefore represent the upper bound for the scores obtainable when the transcripts are automatically segmented.\footnote{It is possible that these upper bounds are exceeded in cases where the reference segmentations of manual transcripts and reference translations are not perfectly aligned, anomalies that we have indeed observed in our benchmarks.} The minimal differences observed between the manual and automatic segmentation scores demonstrate the ability of our segmenters to reconstruct the manual segmentations, to the extent discussed below.

\newpage

\noindent
{\bf Key observations and takeaways.} 

\smallskip

\noindent
{\it -- XL-Segmenter approaches manual segmentation.} The XL-Segmenter works well, as the observed relative degradation compared to manual segmentation is limited to 1-2\%. This result represents a 
baseline for the cross-lingual re-segmentation problem, as XL-Segmenter is a direct extension of L-Segmenter towards its application in cross-lingual settings.

\smallskip
\noindent
{\it -- XLR-Segmenter close the gap with manual segmentation.} The proposed boundary refinement stage allows to close the gap with  manual segmentation, showing consistent improvement over the XL-Segmenter baseline. The use of SimAlign in the refinement step yields slightly better results than LaBSE.

\smallskip
\noindent
{\it -- Our automatic re-segmentation is effective and robust to variable BT quality.} Using a higher quality BT, the one guaranteed by MADLAD compared to NLLB (see Table~\ref{tab:BT-results}), yields only a slightly better segmentation, thus demonstrating the robustness of our re-segmentation procedure in relation to the quality of the BT step it relies on.

\smallskip

These experiments indicate that XL-Segmenter, the direct extension of L-Segmenter for its use across different languages, is extremely effective, and that our boundary refinement stage allows to substantially eliminate all remaining alignment errors.

\begin{table}[ht]
\caption{LASER scores computed on the manually segmented reference translations and the reference transcripts segmented: manually, re-segmented by either XL-Segmenter, XLR-SimAlign or XLR-LaBSE with respect to the BT by either MADLAD or NLLB}
\label{tab:in-vitroReseg}
\centering
\begin{tabular}{lll|cc}
\toprule
                      && segmentation      & MuST-C & Europarl-ST \\
\cline{3-5}
                      && manual   & 0.8550 & 0.9008 \\
\hline
\multirow{6}{*}{BT for reseg} & \multirow{3}{*}{mdld} & XL-Segmenter    & 0.8453 & 0.8847 \\
                      && XLR-SimAlign & 0.8536 & 0.9009 \\
                      && XLR-LaBSE &  0.8519 & 0.8997\\
\cdashline{2-5}
&\multirow{3}{*}{nllb} & XL-Segmenter   & 0.8409 & 0.8757 \\
                      && XLR-SimAlign & 0.8525 & 0.8997\\
                      && XLR-LaBSE &  0.8524 & 0.8991\\
\bottomrule
\end{tabular}
\end{table}

\smallskip

\subsection{In-the-Wild Evaluation}
\label{sec:realistic}

\paragraph{\bf Outline of experiments} Our final set of experiments focuses on realistic working conditions, where audio is automatically segmented, and transcripts are generated by ASR systems. We first evaluate the quality of the source segmentation similarly to what we did under controlled conditions (Section~\ref{sec:reseg-controlled}), with the difference that an automatic segmentation of the audio is used instead of the manual one. The audio is segmented by means of SHAS~\cite{tsiamas22_interspeech}, using the specific model for those languages for which it is available (English, Spanish, French, Italian, and Portuguese), and the multilingual model for the other languages. Each segment is then transcribed with the three ASR systems listed in Section~\ref{sec:ASR-BT}. Finally, automatic transcripts are re-segmented by either XL-Segmenter, XLR-SimAlign, or XLR-LaBSE, exploiting the BT of the reference translations provided by the two BT models also listed in Section~\ref{sec:ASR-BT}. The evaluation of re-segmented automatic transcripts is done both in terms of WER against the manual transcripts and with LASER against the reference translations. Then, following the same methodology adopted under controlled conditions (Section~\ref{sec:exp-src-controlled}), we compare BT sources with those obtained on {\it automatic transcripts} of {\it automatically segmented audio}. This setup enables a comprehensive assessment of the advantages and limitations of both families of synthetic sources when applied in the wild.

\begin{figure}[ht]
\begin{center}
\includegraphics[width=1.0\textwidth]{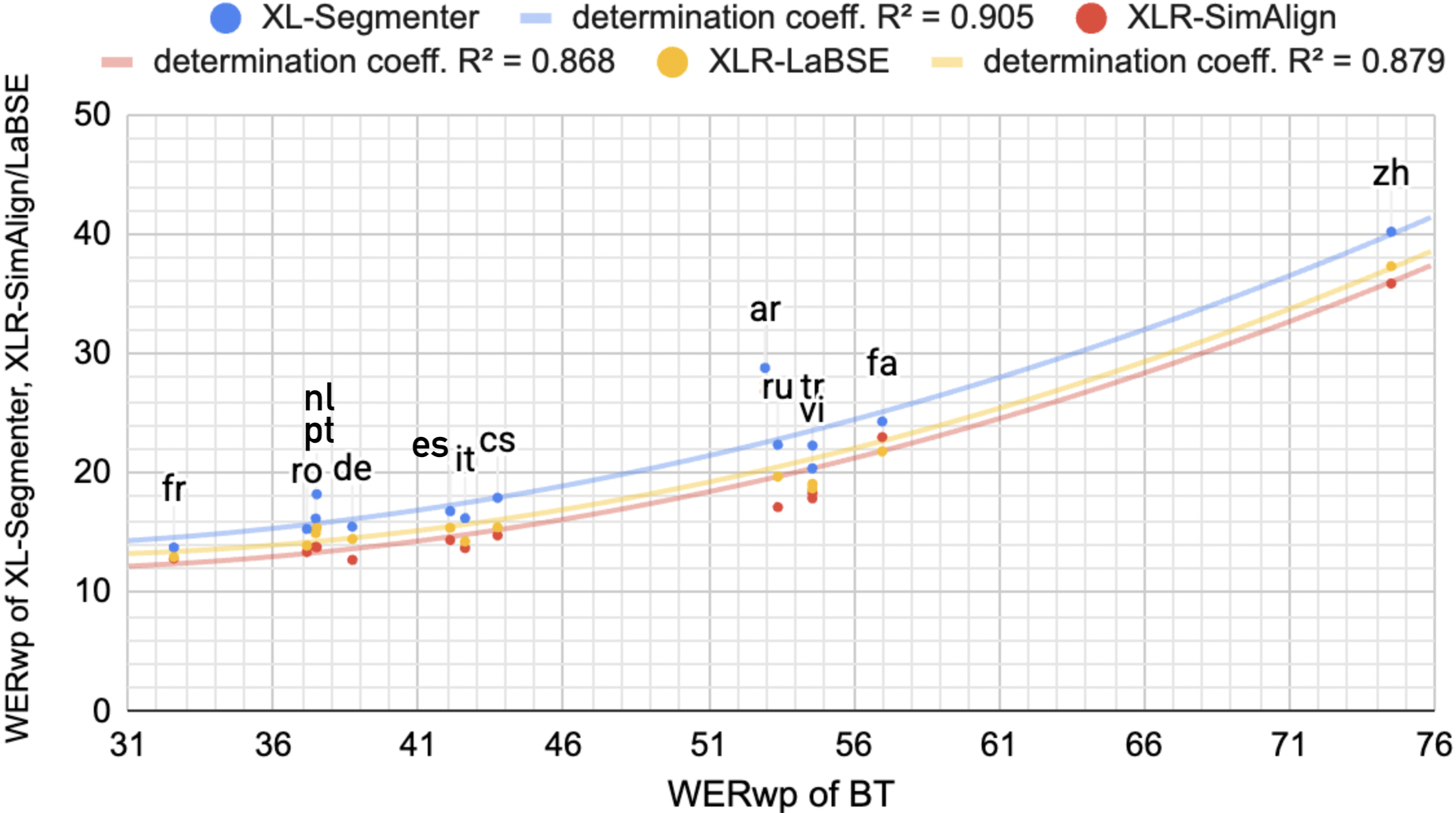}
\end{center}
\caption{MuST-C ${\tt WER_{wp}}$ scores of the automatic re-segmentations shown in Table~\ref{tab:in-vivoReseg-SplitPerLang} plotted against the corresponding ${\tt WER_{wp}}$ of the madlad BT. Quadratic polynomial trend lines are also shown, while coefficients of determination ($\tt R^2$) are given in the legend}
\label{fig:plotWER-CLvsBT}
\end{figure}

\paragraph{\bf In-the-wild Re-Segmentation: Results}
Table~\ref{tab:in-vivoReseg} shows the quality of ASR  sources in terms of LASER and WER. The scores obtained by automatically transcribing the manual audio segments are also provided (rows {\tt manual audio seg}); these serve as an upper bound for the re-segmentation quality. The WER is computed in both ``case-insensitive, no-punctuation'' mode (the standard default) and ``case-insensitive, with-punctuation'' mode, to highlight the ability of our refinement stage to also correct the  placement of punctuation marks. Table~\ref{tab:in-vivoReseg-SplitPerLang}, complemented with Figure~\ref{fig:plotWER-CLvsBT} and Table~\ref{tab:bonoboExample},  focuses on MuST-C ${\tt WER_{wp}}$ of ASR-by-whisper/BT-by-madlad as a representative subset of the experiments reported in aggregate form in Table~\ref{tab:in-vivoReseg}, breaking down its results across all involved language pairs; a per-language breakdown for all configurations would be impractical to present, and the trends observed here are consistent with those of all other settings.

\begin{table}[ht]
\caption{Evaluation of automatically re-segmented automatic transcripts. LASER scores are calculated with respect to the reference translation, and WER scores with respect to the reference transcript. WER is computed ignoring casing and considering ({\tt wp}) or not ({\tt np}) punctuation marks; in {\tt wp} mode, text is tokenized, i.e., words are separated from punctuation. The rows labelled {\tt manual seg} report WER and LASER scores obtained by transcribing the manually segmented audio with each ASR system, without any subsequent re-segmentation step. The LASER score for the manually segmented reference transcript is 0.8550 for MuST-C and 0.9008 for Europarl-ST}
\label{tab:in-vivoReseg}
\setlength{\tabcolsep}{4pt}
\begin{tabular}{llr|ccccccc}
\hline
&& \multicolumn{1}{r}{ASR by $\rightarrow$} & \multicolumn{2}{c}{whisper} & \multicolumn{2}{c}{owsm} & \multicolumn{2}{c}{seamless} \\
&\multicolumn{2}{r}{BT for reseg by $\rightarrow$}  & madlad & nllb & madlad & nllb & madlad & nllb \\
\hline

\multirow{12}{*}{\rotatebox[origin=c]{90}{MuST-C}} &\multirow{4}{*}{\rotatebox[origin=c]{90}{LASER}} & manual audio seg & \multicolumn{2}{c}{0.8256} & \multicolumn{2}{c}{0.8176} & \multicolumn{2}{c}{0.8060}\\  
\cdashline{3-9}
&                    & shas+XL-Segmenter          & 0.8229 & 0.8194 & 0.8162 & 0.8120 & 0.7286 & 0.7253 \\
&                    & shas+XLR-SimAlign          & 0.8292 & 0.8283 & 0.8229 & 0.8217 & 0.7283 & 0.7274 \\
&                    & shas+XLR-LaBSE             & 0.8308 & 0.8307 & 0.8232 & 0.8227 & 0.7315 & 0.7312 \\
\cline{2-9}
&\multirow{4}{*}{\rotatebox[origin=l]{90}{WER}$_{\tt  wp}$}   & manual audio seg & \multicolumn{2}{c}{12.21} & \multicolumn{2}{c}{13.66} & \multicolumn{2}{c}{22.01}\\
\cdashline{3-9}
&                    & shas+XL-Segmenter          & 20.60 & 21.47 & 22.49 & 23.39 & 48.07 & 48.76 \\
&                    & shas+XLR-SimAlign          & 17.76 & 18.05 & 19.64 & 20.06 & 48.04 & 48.37 \\
&                    & shas+XLR-LaBSE             & 18.65 & 18.84 & 20.84 & 21.08 & 51.03 & 51.20 \\
\cline{2-9}
&\multirow{4}{*}{\rotatebox[origin=l]{90}{WER}$_{\tt  np}$}   & manual audio seg & \multicolumn{2}{c}{6.97} & \multicolumn{2}{c}{9.37} & \multicolumn{2}{c}{18.53}\\
\cdashline{3-9}
&                    & shas+XL-Segmenter          & 14.46 & 15.49 & 16.95 & 17.95 & 43.83 & 44.58 \\
&                    & shas+XLR-SimAlign          & 11.61 & 11.93 & 14.02 & 14.47 & 43.66 & 44.03 \\
&                    & shas+XLR-LaBSE             & 12.73 & 12.94 & 15.43 & 15.70 & 46.76 & 46.95 \\
\hline
\multirow{12}{*}{\rotatebox[origin=c]{90}{Europarl-ST}} &\multirow{4}{*}{\rotatebox[origin=c]{90}{LASER}} & manual audio seg & \multicolumn{2}{c}{0.8849} & \multicolumn{2}{c}{0.8604} & \multicolumn{2}{c}{0.8878}\\  
\cdashline{3-9}
&                    & shas+XL-Segmenter          & 0.8662 & 0.8575 & 0.8375 & 0.8280 & 0.8667 & 0.8579\\
&                    & shas+XLR-SimAlign          & 0.8830 & 0.8818 & 0.8526 & 0.8509 & 0.8811 & 0.8799 \\
&                    & shas+XLR-LaBSE             & 0.8830 & 0.8823 & 0.8534 & 0.8524 & 0.8819 & 0.8810 \\
\cline{2-9}
&\multirow{4}{*}{\rotatebox[origin=l]{90}{WER}$_{\tt  wp}$}   & manual audio seg & \multicolumn{2}{c}{14.61} &\multicolumn{2}{c}{24.52} & \multicolumn{2}{c}{13.43}\\
\cdashline{3-9}
&                    & shas+XL-Segmenter          & 23.35 & 26.13 & 34.17 & 37.11 & 23.21 & 25.98 \\
&                    & shas+XLR-SimAlign          & 18.76 & 19.11 & 30.20 & 30.69 & 19.02 & 19.37 \\
&                    & shas+XLR-LaBSE             & 19.07 & 19.28  & 30.57  & 30.86  & 19.30  & 19.54 \\
\cline{2-9}
&\multirow{4}{*}{\rotatebox[origin=l]{90}{WER}$_{\tt  np}$}   & manual audio seg & \multicolumn{2}{c}{10.40} &\multicolumn{2}{c}{20.71} & \multicolumn{2}{c}{9.84}\\
\cdashline{3-9}
&                    & shas+XL-Segmenter          & 17.99 & 20.92 & 29.07 & 32.19 & 18.08 & 20.98 \\
&                    & shas+XLR-SimAlign          & 13.49 & 13.84 & 25.02 & 25.51 & 13.85 & 14.20 \\
&                    & shas+XLR-LaBSE             & 13.97 & 14.19  & 25.66  & 25.97  & 14.34  & 14.60 \\
\hline
\end{tabular}
\end{table}

\bigskip

\noindent
{\bf In-the-wild Re-Segmentation: Key observations and takeaways.}

\smallskip
\noindent
{\it -- The boundaries refiner handles real-world conditions.} The refinement stage proves to be extremely effective also in realistic conditions: both XLR-SimAlign and XLR-LaBSE consistently outperform XL-Segmenter, independently of the ASR system (Whisper, Owsm, Seamless), the corpus (MuST-C, Europarl-ST), and the metric (LASER, WER$_{\tt wp}$, WER$_{\tt np}$). The observed WER improvement, up to more than 7 absolute points (from WER$_{\tt np}$=20.92 to 13.84 for Whisper/NLLB on Europarl-ST), confirms the ability of the refinement stage to correctly reposition words misplaced by XL-Segmenter.

\smallskip
\noindent
{\it -- The semantics of automatic transcripts is fully preserved.} In terms of LASER, the re-segmentation of automatic transcripts enables approaching the manual segmentation of automatic transcripts on Europarl-ST (e.g., on Whisper/MADLAD: 0.8830 vs. 0.8849) and even improves it on MuST-C (e.g., on Whisper/MADLAD: 0.8292/0.8308 vs. 0.8256). This means that, despite the presence of transcription errors, both XLR-SimAlign and XLR-LaBSE can adjust the source text segmentation inherited by the SHAS-based audio segmentation in such a way as to recover the semantic correspondence between the source and target segments.

\smallskip

\noindent
{\it -- The semantics of manual transcripts is almost fully preserved.}  The fully automated process is able to generate segmented ASR texts (with {\tt whisper}), aligned to the reference translations, whose LASER scores are 0.8292/0.8308  for MuST-C and 0.8830 for Europarl-ST. These are only 2-3\% worse than the upper bound scores obtained by computing LASER against manually segmented reference transcripts (0.8550 for MuST-C and 0.9008 for Europarl-ST, see Table~\ref{tab:in-vitroReseg}). This further demonstrates the effectiveness of our source re-segmentation solution.

\smallskip
\noindent
{\it -- A few words are still misplaced.}  In terms of WER, the re-segmentation by both XLR-SimAlign and XLR-LaBSE of automatic transcripts generated on the SHAS audio segmentation approaches the manual segmentation on Europarl-ST across all ASR systems, and equally so on MuST-C for Whisper and Owsm ASRs; the approximation is less tight for Seamless on MuST-C. The minimum degradation observed across the two corpora amounts to 3.09 ${\tt WER_{np}}$ points on Europarl-ST (ASR$_{\tt whspr-mdld}$, from 10.40 to 13.49) and 4.64 ${\tt WER_{np}}$   points on MuST-C (ASR$_{\tt whspr-mdld}$, from 6.97 to 11.61). The gap is ascribable to the cases in which the re-segmentation does not correctly reposition words of the automatic transcript. Table~\ref{tab:wrongResegExample} illustrates one such case. The two ASR segments shown contain no transcription errors. The issue lies in the phrase ``in your city.'', which is present in the audio recording, as confirmed by its appearance in the reference source transcript, but was omitted in the Italian reference translation and consequently in its BT. Lacking any aligned counterpart in the BT, the re-segmenter cannot determine where to place this phrase: XLR-SimAlign moves it entirely to the beginning of the following segment, including the full stop; XLR-LaBSE splits it, retaining ``in'' at the end of the first segment and shifting ``your city.'' to the second. However, as the LASER results indicate, these errors do not alter the overall semantic content of the segments.

\begin{table}[ht]
\caption{Example of a re-segmentation failure case. The block ASR shows two contiguous automatic transcripts (partially). The two following parallel blocks correspond to two contiguous segments produced by the re-segmentation stage. Each block shows: the reference target translation (ref trg), its back-translation (BT), the reference source transcript (src ref), and the outputs of XLR-SimAlign and XLR-LaBSE.}
\label{tab:wrongResegExample}
\setlength{\tabcolsep}{1.5pt}
\small
\begin{tabular}{l|l}
\hline
ASR & [...] It's generally where you will get if you call 311 in your city.\\
    & If you should ever have the chance [...] \\
\hline
ref trg & È qui che arrivano le vostre chiamate quando digitate il 311. \\
BT      & 		This is where your calls come when you dial 311.\\
src ref		& It's generally where you will get if you call 311 in your city.\\
XLR-SimAlign &	It's generally where you will get if you call 311\\
XLR-LaBSE	& It's generally where you will get if you call 311 in \\
\hline
ref trg &		Se vi capitasse [...] \\
BT  	&	If you happen to be [...] \\
src ref	 &	If you should ever have the chance [...] \\
XLR-SimAlign	& in your city. If you should ever have  the chance [...] \\
XLR-LaBSE	& your city. If you should ever have the chance [...]\\
\hline
\end{tabular}
\end{table}

\smallskip 

\begin{table}[ht]
\caption{MuST-C ${\tt WER_{wp}}$ scores of the manual segmentation and of the automatic re-segmentations by XL-Segmenter, XLR-SimAlign and XLR-LaBSE (based on madlad BT) of the whisper transcription broken down  per target language. The {\tt avg} column reports the aggregate values shown in Table~\ref{tab:in-vivoReseg}}
\label{tab:in-vivoReseg-SplitPerLang}
\setlength{\tabcolsep}{2pt}
\footnotesize
\begin{tabular}{l|cccccccccccccc|c}
\hline
segmentation & ar & cs & de & es & fa & fr & it & nl & pt & ro & ru & tr & vi & zh & avg\\
\hline
man audio seg & 15.78 & 11.87 & 10.84 & 10.97 & 12.68 & 10.86 & 10.90 & 10.80 & 10.98 & 10.96 & 10.99 & 11.28 & 11.98 & 20.12 & 12.21 \\
XL-Segmenter & 28.81 & 17.93 & 15.51 & 16.82 & 24.33 & 13.78 & 16.23 & 18.23 & 16.20 & 15.32 & 22.38 & 22.31 & 20.39 & 40.21 & 20.60 \\
XLR-SimAlign & 26.86 & 14.79 & 12.73 & 14.39 & 23.02 & 12.85 & 13.73 & 13.78 & 13.83 & 13.39 & 17.16 & 18.34 & 17.89 & 35.89 & 17.76 \\
XLR-LaBSE & 27.55 & 15.46 & 14.49 & 15.42 & 21.81 & 12.97 & 14.27 & 15.41 & 14.98 & 13.92 & 19.70 & 19.10 & 18.72 & 37.34 & 18.65 \\
\hline
\end{tabular}
\end{table}

\begin{table}[ht]
\caption{Example of BT quality from Chinese and from French. ${\tt WER_{wp}}$ is computed on shown texts, while for MetricX the text is first detokenized. Reference Chinese and French texts used by MetricX as source input are not shown}
\label{tab:bonoboExample}
\setlength{\tabcolsep}{1.5pt}
\small
\begin{tabular}{l|l|cc}
\hline
source & text & ${\tt WER_{wp}\hspace{-1mm}\downarrow}$ & MetricX$\downarrow$ \\
\hline
ref src& bonobos are , together with chimpanzees , your living closest relative .	& - & - \\
BT zh	 & bonobos and chimpanzees are the closest relatives in our lives .		& 83.33 & 1.398 \\
BT fr	 & bonobos are , along with chimpanzees , your closest living cousins .	& 33.33 & 1.612 \\
\hline
\end{tabular}
\end{table}

\noindent
{\it --  Per-language analysis.} 
Table~\ref{tab:in-vivoReseg-SplitPerLang} breaks down the ${\tt WER_{wp}}$ aggregate scores of Table~\ref{tab:in-vivoReseg}  across all 14 target languages for a representative configuration i.e., Whisper/MADLAD on MuST-C. All the other configurations behave in an analogous way.The per-language results fully confirm the aggregate trends: the refinement stage consistently reduces WER with respect to XL-Segmenter across all languages, regardless of whether XLR-SimAlign or XLR-LaBSE is used. XLR-SimAlign yields lower WER than XLR-LaBSE in all but one case, Farsi, where LaBSE achieves 21.81 against SimAlign's 23.02, an isolated exception that does not alter the overall picture. Figure~\ref{fig:plotWER-CLvsBT} reveals a further insight: there is a quadratic relationship between the ${\tt WER_{wp}}$ of the re-segmented transcripts (y-axis) and the ${\tt WER_{wp}}$ of the BT on which the re-segmentation is built (x-axis), with $\tt R^2$ values ranging from 0.868 to 0.905. This suggests that the higher the lexical divergence between the BT and the reference, as measured by WER, the harder the re-segmenter's task. This lexical divergence does not, however, imply semantic divergence, as Table~\ref{tab:bonoboExample} demonstrates for the two language pairs at opposite extremes in Figure~\ref{fig:plotWER-CLvsBT}, {\tt en-fr} and {\tt en-zh}. Despite a ${\tt WER_{wp}}$ of 83.33 for the Chinese BT against only 33.33 for the French one, the MetricX score of the Chinese BT (1.398) is actually better than that of the French BT (1.612), confirming that the high WER of the Chinese BT reflects lexical and syntactic divergence from the reference rather than a semantic one, and that the BTs used in our experiments are of adequate quality.

\smallskip

These  in-the-wild re-segmentation experiments show that, even under realistic conditions where ASR sources are generated fully automatically, our proposed alignment solution remains highly effective, yielding only a minimal degradation in semantic content, as measured by LASER, of 2–3\% compared to the upper bound scores computed on manual sources.

In light of results shown in Tables~\ref{tab:in-vivoReseg} and~\ref{tab:in-vivoReseg-SplitPerLang} and discussed above, all subsequent experiments use XLR-SimAlign for the refinement stage, as it slightly outperforms XLR-LaBSE overall. Although XLR-LaBSE is substantially faster, as discussed in Appendix~B, we opt for accuracy over computational efficiency.

\begin{table}[ht!]
\caption{Pearson correlations between the MetricX scores of each system on MuST-C, computed (i) by using as the source the reference transcript (i.e., in the standard way), and (ii) by using as the source the automatically re-segmented automatic transcripts generated from automatic audio segmentation. Each reported correlation coefficient is the average of the correlation coefficients computed separately for each language pair. For \textdagger~(correlations potentially biased) and heat map colouring see caption of Table~\ref{tab:corr-mustc}}

\label{tab:corr-mustc-shas}
\setlength{\tabcolsep}{2pt}
\includegraphics[width=1.0\textwidth]{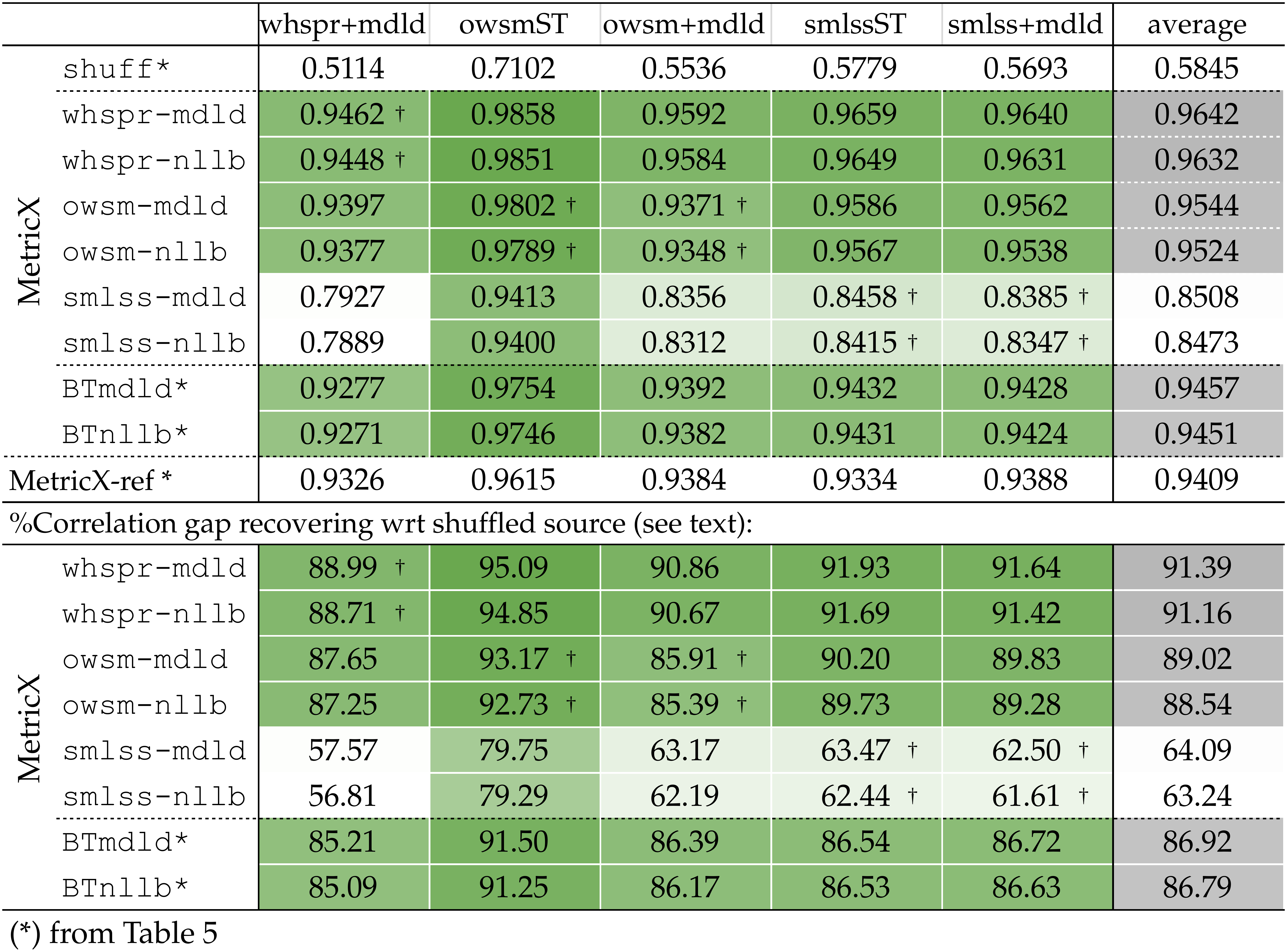}
\end{table}

\begin{table}[ht]
\caption{Pearson correlations between the MetricX scores of each system on Europarl-ST, computed (i) by using as the source the reference transcript (i.e. in the standard way), and (ii) by using as the source the automatically re-segmented automatic transcripts generated from automatic audio segmentation. Each reported correlation coefficient is the average of the correlation coefficients computed separately for each language pair. For \textdagger~(correlations potentially biased) and heat map colouring see caption of Table~\ref{tab:corr-mustc}}

\label{tab:corr-epst-shas}
\setlength{\tabcolsep}{2pt}
\includegraphics[width=1.0\textwidth]{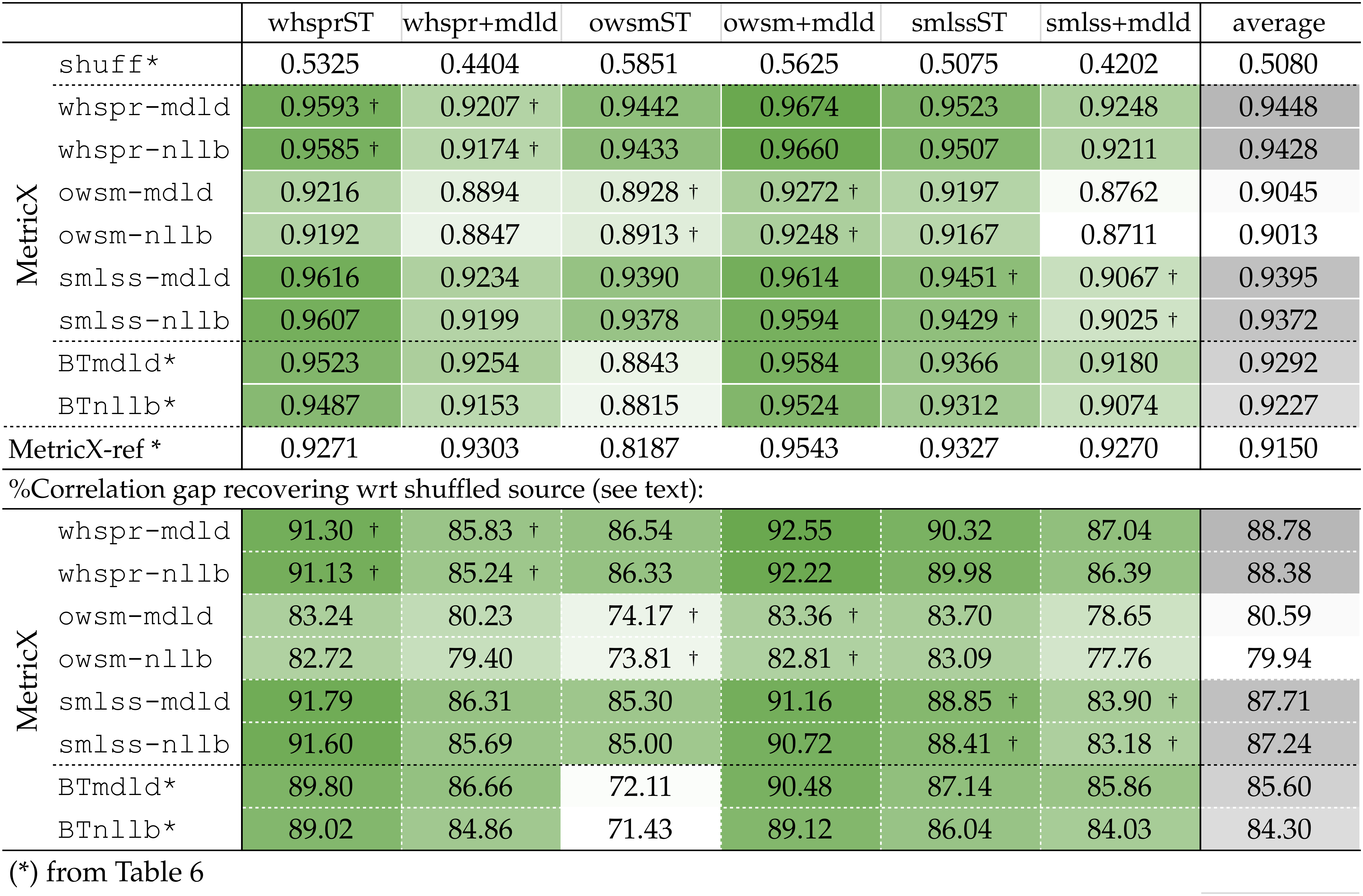}
\end{table}

\begin{figure}[ht]
\begin{center}
\includegraphics[width=0.5\textwidth]{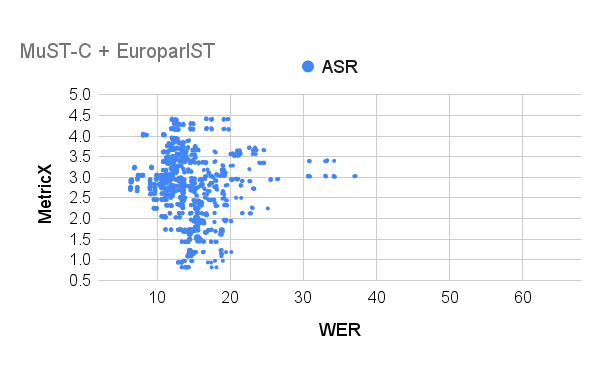}\includegraphics[width=0.5\textwidth]{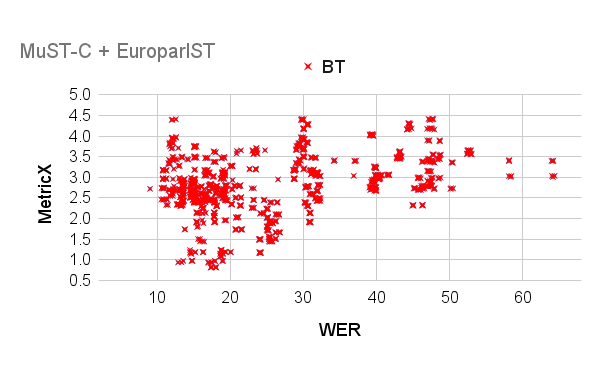}
\end{center}
\caption{The plane of these  scatter charts is defined by the WER and MetricX scores of the ASR and BT sources, respectively. For all possible comparisons between the MetricX correlation with the ASR source and that with the BT source, computed on all language pairs of the two corpora and for all ST systems, the two charts show where the cases in which it was preferable to use the ASR (on the left) or the BT (on the right) as the source for the computation of MetricX are placed in that plane. Biased ASR MetricXs, i.e. those of ST systems that are somehow involved in the generation of the ASR source of the metric, are excluded. The total number of points is 3440, 2104  on the left (ASR wins, 61.2\%), 1336 on the right (BT wins, 38.8\%). A random 1\% change was applied to all values in order to avoid the overlapping of points and make all of them visible.} 
\label{fig:plotUnion-realistic}
\end{figure}

\begin{figure}[ht]
\begin{center}
\includegraphics[width=0.5\textwidth]{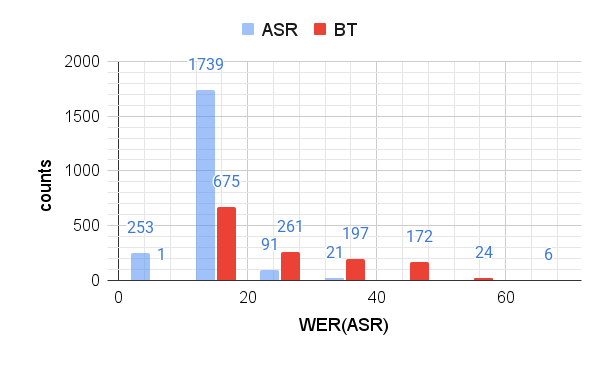}\includegraphics[width=0.5\textwidth]{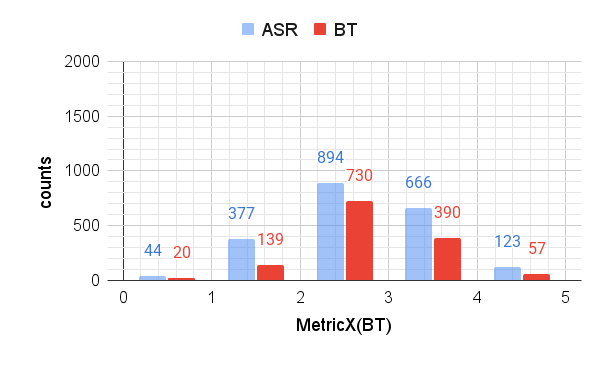}
\end{center}
\caption{For all possible comparisons between the MetricX correlation with the ASR source and that with the BT source, computed on all language pairs of the two corpora and for all ST systems, these histograms illustrate the distribution of cases in which the standard MetricX shows a higher correlation with MetricX using either the ASR or the BT as source input. Biased ASR MetricXs, i.e. those of ST systems that are somehow involved in the generation of the ASR source of the metric, are excluded. The left chart reflects this distribution as a function of transcription quality (WER), while the right chart does so with respect to (back-)translation quality (MetricX).}
\label{fig:histoUnion-realistic}
\end{figure}

\smallskip

\begin{table}[ht]
\caption{For all possible comparisons between the MetricX correlation with the ASR source and that with the BT source, computed on all language pairs of the two corpora and for all ST systems, the total number of wins per synthetic source type (ASR - whisper, owsm, seamless resegmented by XLR-SimAlign wrt madlad or nllb - or BT - madlad, nllb) is given here. Biased ASR MetricXs, i.e., those of ST systems that are somehow involved in the generation of the ASR source of the metric, are excluded}
\label{tab:winners-union-realistic}
\setlength{\tabcolsep}{1.7pt}
\begin{tabular}{l|cccc|cccc|cccc}
\hline
\multirow{3}{*}{corpus} 
& \multicolumn{4}{c|}{WER$_{\tt ASR}\le$20\%} & \multicolumn{4}{c|}{WER$_{\tt ASR}>$20\%} & \multicolumn{4}{c}{total} \\
& \multicolumn{2}{c}{ASR wins} & \multicolumn{2}{c|}{BT wins}& \multicolumn{2}{c}{ASR wins} & \multicolumn{2}{c|}{BT wins}& \multicolumn{2}{c}{ASR wins} & \multicolumn{2}{c}{BT wins} \\
& counts & \% & counts & \% & counts & \% & counts & \% & counts & \% & counts & \% \\
\hline
MuST-C       & 335  & 99.7 & 1   & 0.3  & 38  & 17.0 &  186 & 83.0 & 373  & 66.6 & 187 & 33.4 \\
Europarl-ST  & 1657 & 71.1 & 675 & 28.9 & 74  & 13.5 &  474 & 86.5 & 1731 & 60.1 & 1149 & 39.9 \\
\hline
total        & 1992 & 74.7 & 676 & 25.3 & 112 & 14.5 &  660 & 85.5 & 2104 & 61.2 & 1336 & 38.8 \\
\hline

\end{tabular}
\end{table}

\paragraph{\bf In-the-wild Synthetic Sources: Results}

Tables~\ref{tab:corr-mustc-shas} and~\ref{tab:corr-epst-shas} provide, for the two corpora, the correlations achieved by the various  ASR sources generated in realistic conditions, together with the corresponding gap recovery with respect to the shuffled sources. The ASR sources are those produced by our two-stage re-segmentation algorithm, whose quality is shown in rows {\tt shas+XLR-SimAlign} of Table~\ref{tab:in-vivoReseg}.  Due to the observed lower sensitivity of COMET to the source content (see Section~\ref{sec:exp-src-controlled}), here we focus on MetricX, omitting COMET results (which are reported in Appendix~E for completeness). It is worth noting that realistic conditions do not impact the  BT sources, as they are generated by back-translating segment by segment the reference translations. For ease of reference, MetricX$_{\tt BTmdld}$ and MetricX$_{\tt BTnllb}$ scores that are reported in Tables~\ref{tab:corr-mustc-shas} and~\ref{tab:corr-epst-shas} correspond to those originally reported in Tables~\ref{tab:corr-mustc} and~\ref{tab:corr-epst}. Figure~\ref{fig:plotUnion-realistic} and Figure~\ref{fig:histoUnion-realistic} are equivalent to Figures~\ref{fig:plotUnion} and~\ref{fig:histoUnion}, but related to realistic conditions under investigation in this last set of experiments.

\newpage
\noindent
{\bf In-the-wild Synthetic Sources: Key observations and takeaways.}

\smallskip
\noindent
{\it -- ASR sources remain effective.} 
In general, compared with those obtained in controlled conditions, the correlations of ASR-derived sources marginally decrease, but remain consistently above 0.92. The few exceptions concern the ASR sources derived from high-WER transcripts, namely, those generated by SeamlessM4T on MuST-C and by OWSM on Europarl-ST. Even in these cases, however, values are always above 0.80 (with a minimum value of 0.8140), which still reflects a strong correlation. These results indicate that, even under realistic conditions,  ASR sources remain effective substitutes for manual transcripts.

\smallskip
\noindent
{\it -- ASR confirms to be a better proxy.} 
In 9 cases out of 11 (for 5/5 ST systems on MuST-C and for 4/6 ST systems on Europarl-ST) there is at least one ASR source that is more effective than any BT source, indicating a generic greater effectiveness of ASR sources compared to BT sources also in this challenging, realistic setting.

\smallskip
\noindent
{\it -- The 20\% threshold is still discriminative.} 
Figures~\ref{fig:plotUnion-realistic} and~\ref{fig:histoUnion-realistic} remain overall similar to those obtained under controlled conditions, thus confirming our findings in Section~\ref{sec:exp-src-controlled} and, in particular, that (i)  ASR sources are generally more effective than the BT ones, and (ii) the 20\% WER threshold discriminates between them. These findings are also supported by the results reported in Table~\ref{tab:winners-union-realistic}, not too dissimilar from those in Table~\ref{tab:winners-union-controlled}. It emerges that ASR is preferable in 61.2\% of all cases, but also that this preference rises to 74.7\% when the WER is at most 20\%. Conversely, when the WER exceeds 20\%, BT is the best choice in 85.5\% of cases. 

\begin{table}[ht]
\caption{Per-language MetricX scores and correlations for the owsmST system on MuST-C. The aggregate values of correlations are reported in Table~\ref{tab:corr-mustc-shas} as well, while the aggregate value of the BLEU score is also reported in Table~\ref{tab:ST-results}. (*) The BLEU score for Chinese is at character level.  See text for further details.}
\label{tab:corr-mustc-shas-SplitPerLang}
\setlength{\tabcolsep}{1.6pt}
\footnotesize
\begin{tabular}{l|cccccccccccccc|c}
\hline
{\tt owsmST} & ar & cs & de & es & fa & fr & it & nl & pt & ro & ru & tr & vi & zh & avg\\
\hline
BLEU & 8.94 & 7.75 & 26.99 & 22.08 & 9.19 & 27.29 & 17.49 & 22.02 & 21.91 & 14.92 & 10.05 & 7.15 & 14.50 & 16.27* & 16.18 \\
\hline
{\tt r}({\tt whspr-mdld}) & .9827 &  .9864 &  .9816 &  .9878 &  .9783 &  .9909 &  .9911 &  .9887 &  .9881 &  .9897 &  .9901 &  .9807 &  .9920 &  .9725 &  .9858 \\
{\tt r}({\tt BTmdld}) & .9771 & .9797 & .9682 & .9753 & .9677 & .9857 & .9765 & .9834 & .9770 & .9776 & .9725 & .9724 & .9778 & .9644 & .9754 \\ 
{\tt r}(MetricX-ref) & .9639 &  .9545 &  .9556 &  .9614 &  .9550 &  .9625 &  .9624 &  .9635 &  .9642 &  .9640 &  .9636 &  .9583 &  .9697 &  .9625 &  .9615 \\
\hline
\end{tabular}
\end{table}

\noindent
{\it --  Per-language analysis.} 
Table~\ref{tab:corr-mustc-shas-SplitPerLang} breaks down, for two specific set of experiments involving the {\tt owsmST} system on MuST-C, the aggregate results of Table~\ref{tab:corr-mustc-shas}, across all 14 target languages. The first row reports the quality of the translations produced by the system on the test set, as measured by BLEU. The second and third rows ({\tt r(whspr-mdld), r(BTmdld)}) show, for each language-pair section of the test set, the correlation between standard MetricX scores and those of the synthetic MetricX variant using the {\tt whspr-mdld} and {\tt BTmdld} sources, respectively; the averages (0.9858, 0.9754) correspond to the values reported in the relevant cells of Table~\ref{tab:corr-mustc-shas}. The fourth row reports the correlation between MetricX-ref, the source-free variant of the metric, and the standard MetricX, with an average of 0.9615, again corresponding to the relevant cell of Table~\ref{tab:corr-mustc-shas}. Against a backdrop of widely varying and generally low translation quality (BLEU scores ranging from 7.15 for {\tt tr} to 27.29 for {\tt fr}, with an average of 16.18, consistent with the low quality already observed in Table~\ref{tab:ST-results}), correlation coefficients remain uniformly high and appear to be independent of the quality of the translations under evaluation. Moreover, our synthetic MetricX variants consistently outperforms the source-free version across all language pairs.

\smallskip
These final experiments demonstrate that, even under realistic conditions, synthetic MetricX maintains a strong correlation with its standard version based on manual transcripts. Furthermore, they also confirm that, when transcripts are of sufficient quality (WER not exceeding 20\%), ASR is the preferred synthetic source,  whereas for higher WERs, BT provides a better alternative.

\subsection{Low-Resource Evaluation}
\label{sec:additional-exps_low-resource}

The experiments presented so far were conducted on two corpora characterized by substantial linguistic diversity and challenging translation directions. However, the language pairs covered by both datasets benefit from abundant resources and, crucially, are supported by the underlying models employed by the evaluation metrics under investigation. A natural question therefore arises: how do COMET or MetricX behave in low-resource settings, particularly for language pairs that are not explicitly covered by the two metrics?  This issue is addressed in the following, where results on the Bemba-English language pair are presented and discussed.

\newpage
\noindent
{\bf Evaluation on a Low-Resource Language Pair: Bemba-English}

\noindent
In recent years, the annual evaluation campaign on spoken language translation organized within IWSLT\footnote{\url{https://iwslt.org}} has included a shared task on speech-to-text translation in low-resource conditions. In the 2025 edition, one of the proposed translation directions was Bemba-English. Bemba is a Bantu language spoken by over 10 million people in Zambia and other parts of Africa. The public availability of the IWSLT 2025 test set as part of a larger corpus~\cite{sikasote-etal-2023-big} makes this task particularly suitable for the purposes of the present investigation. In addition, the organizers kindly granted us access to the primary runs submitted by the participating teams. Finally, several transcription and translation models adapted to Bemba are openly available. Among these, we selected the following ones based on the performance figures reported by their respective authors, with the aim of covering a broad range of quality levels, from strong to weak, so as to ensure a diverse and informative experimental setup:

\smallskip

\noindent
{\bf ASR whsprAda}:\footnote{\url{https://huggingface.co/kreasof-ai/whisper-medium-bem2eng}} fine-tuned version of the Whisper {\tt medium} model on Bemba speech.

\noindent
{\bf ASR mms}:\footnote{\url{https://huggingface.co/facebook/mms-1b-all}} fine-tuned version on 1162 languages of the 1-billion-parameters  Massively Multilingual Speech model.

\noindent
{\bf ASR mmsAda}:\footnote{\url{https://huggingface.co/csikasote/mms-1b-all-bemgen-combined-sd-1e-1}} fine-tuned version on Bemba speech data of the  {\tt mms} model.

\noindent
{\bf MT nllb3.3bAda}:\footnote{\url{https://huggingface.co/kreasof-ai/nllb-200-3.3B-bem2eng-bigc-flores200-tatoeba}} fine tuned version on Bemba-English parallel textual data of the 3.3B parameters NLLB model.

\noindent
{\bf MT nllb600mAda}:\footnote{\url{https://huggingface.co/kreasof-ai/nllb-200-600M-eng2bem}} fine tuned version on English-Bemba parallel textual data of the distilled-600 million parameters NLLB model.

\noindent
{\bf MT hlsnk}:\footnote{\url{https://huggingface.co/Helsinki-NLP/opus-mt-en-bem}} MT model of the Helsinki-NLP/opus-mt family specialized for the translation from English to Bemba.

\smallskip

The ST systems involved in this investigation are:

\smallskip
\noindent
{\bf mmsAda+nllb3.3bAda}:
cascade of the {\tt ASR mmsAda} and {\tt MT nllb3.3bAda} models.

\noindent
{\bf JHU@iwslt2025}: primary run of the Johns Hopkins University submission to the Bemba-to-English ST task at IWSLT 2025.

\noindent
{\bf KIT@iwslt2025}: primary run of the Karlsruhe Institute of Technology submission to the Bemba-to-English ST task at IWSLT 2025.

\smallskip

\begin{table}[ht]
\small
\caption{Performance of ASR and MT models, and of ST systems on the IWSLT25 Bemba-English test set}
\label{tab:bembaST}
\setlength{\tabcolsep}{2pt}
\begin{tabular}{l|c|c|cccc}
\toprule
\multirow{1}{*}{model/system} & \multirow{1}{*}{used for } & \multirow{1}{*}{lang} & WER\%$\downarrow$  & CER\%$\downarrow$  \\
\midrule
ASR whsprAda   & ASR                   & bem & 31.57 & 9.82 \\
ASR mmsAda     & 
ASR$_{\tt mmsAda+nllb3.3bAda}$
& bem & 64.78 & 16.08 \\
ASR mms        & ASR                   & bem & 73.40 & 17.24 \\
\midrule
\midrule
\multirow{2}{*}{model/system} & \multirow{2}{*}{used for} & lang & BLEU & BLEU-char & COMET & MetricX \\
&& pair & (0-100$\uparrow$) & (0-100$\uparrow$) & (0-1$\uparrow$) & (0-25$\downarrow$) \\
\midrule
MT nllb3.3bAda & 
MT$_{\tt mmsAda+nllb3.3bAda}$
& bem$\rightarrow$en & 29.59 & 59.00 & .7287 & 7.526\\
MT nllb600mAda & BT        & en$\rightarrow$bem            & 8.82  & 55.53 & .6524 & 8.732 \\
MT hlsnk       & BT        & en$\rightarrow$bem            & 1.69  & 38.28 & .5883 & 10.219 \\
\hdashline
mmsAda+nllb3.3bAda &  ST        & bem$\rightarrow$en            & 25.27 & 54.55 & .6961 & 8.437 \\
JHU@iwslt2025 &  ST        & bem$\rightarrow$en            & 30.37 & 58.56 & .7185 & 7.911 \\
KIT@iwslt2025 &  ST        & bem$\rightarrow$en            & 30.78 & 58.61 & .7171 & 7.831 \\
\bottomrule
\end{tabular}
\end{table}

Table~\ref{tab:bembaST} shows the performance on the IWSLT25 Bemba-English test set of these models and systems.\footnote{BLEU scores are computed with sacreBLEU (signature: \texttt{nrefs:1|case:lc|eff:no|tok:13a|smooth:exp|version:2.5.1}) in case-insensitive, no-punctuation mode. To ensure comparability across all systems examined here, we recomputed all scores under identical conditions, rather than relying on those reported in~\cite{agostinelli-etal-2025-findings}. The values may therefore differ from the published ones; one possible source of discrepancy is the handling of apostrophes: here, contractions, possessives, and negative forms (e.g., she's, child's, doesn't) are preserved, a choice that may partially explain the observed differences.} Among the three ASR models, one, {\tt whsprAda}, achieves (relatively) strong performance, whereas the other two exhibit substantially higher WER. However, the actual quality of their Bemba transcripts is better than WER alone might suggest. The percentage gap from {\tt whsprAda} in terms of CER is considerably smaller than the corresponding gap in WER. This discrepancy indicates that the main difficulty lies in adequately word segmenting a character stream that is otherwise reasonably faithful to the underlying acoustic signal. Given that, we selected the transcripts generated by the two models at opposite ends of the performance spectrum as ASR sources for COMET and MetricX, thereby amplifying the potential impact of transcription accuracy on metric behavior. The intermediate ASR model, instead, was used within the  {\tt mmsAda+nllb3.3bAda} cascade system that we compare against the two IWSLT participants.

With respect to the MT models, only one system directly supports translation from Bemba into English. This model achieves solid performance and is employed within the prototype cascade system. The other two models operate in the reverse direction and their transcripts are employed as BT sources for COMET and MetricX; again, they differ substantially in performance, allowing to assess the impact of BT quality on the reliability of the source-based ST evaluation metrics.

The two systems participating at IWSLT 2025, JHU and KIT, exhibit very similar overall performance, with different metrics favoring one or the other. To mitigate the risk of drawing conclusions from nearly indistinguishable scores, the {\tt mmsAda+nllb3.3bAda} cascade system built from fine-tuned ASR and MT components was included in the evaluation. Although its performance is lower than that of the two official systems, it remains competitive enough to provide a meaningful contrast.

Table~\ref{tab:bemba} presents what is the core objective of this section: the behavior of the standard versions and the synthetic variants of COMET and MetricX when evaluating real ST system outputs under low-resource conditions. For each of the three ST systems under investigation, the {\tt r} columns provide the Pearson correlation between each synthetic variant and its corresponding standard metric, the  $\mu$ columns report the mean scores of the evaluation metrics, and the {\tt RMSD} columns indicate the root mean square deviation between each standard value and its synthetic counterpart.

\begin{table}[ht]
\caption{For each ST system, the column $\mu$ provides the mean of the values of the standard and of the synthetic versions of the two source-based evaluation metrics; the column {\tt r}, the Pearson correlation coefficient between the values of the standard version and those of each synthetic variant of the metrics; the column {\tt RMSD}, the Root mean square deviation between the standard and the synthetic values}
\label{tab:bemba}
\setlength{\tabcolsep}{2pt}
\begin{tabular}{ll|ccc|ccc|ccc}
\hline
& \multirow{2}{*}{source}& \multicolumn{3}{c|}{Prototype} & \multicolumn{3}{c|}{JHU@iwslt25} & \multicolumn{3}{c}{KIT@iwslt25} \\
 &  & {\tt r} & $\mu$  & RMSD & {\tt r} & $\mu$  & RMSD & {\tt r} & $\mu$  & RMSD\\
\hline
\multirow{6}{*}{\rotatebox[origin=c]{90}{COMET}} & ref (standard) & & .6961 & & & .7185 & & & .7171 & \\
& ASRwhsprAda     & .9970  & .6962  & 0.922e-2 & .9974 & .7185 & 0.893e-2 & .9968 & .7171 & 0.915e-2 \\
& ASRmms          & .9918  & .7076  & 1.937e-2 & .9925 & .7293 & 1.874e-2 & .9906 & .7286 & 1.953e-2 \\
& BTnllb600mAda   & .9927  & .6973  & 1.451e-2 & .9934 & .7197 & 1.417e-2 & .9921 & .7182 & 1.440e-2 \\
& BThlsnk         & .9914  & .6966  & 1.577e-2 & .9922 & .7191 & 1.541e-2 & .9907 & .7175 & 1.568e-2 \\
& shuf            & .9880  & .6945  & 1.860e-2 & .9898 & .7171 & 1.754e-2 & .9877 & .7152 & 1.805e-2 \\
\hline
\multirow{6}{*}{\rotatebox[origin=c]{90}{MetricX}} & ref (standard) & & 8.437 & & & 7.911 & & & 7.831 & \\
& ASRwhsprAda    & .9659 & 8.398 & 0.938 & .9649 & 7.859 & 0.957 & .9575 & 7.803 & 0.956 \\
& ASRmms         & .7982 & 13.691& 5.726 & .8016 & 13.582 & 6.112 & .7600 & 13.397 & 5.979 \\
& BTnllb600mAda  & .9287 & 8.485 & 1.365 & .9269 & 7.985 & 1.402 & .9129 & 7.872 & 1.385 \\
& BThlsnk        & .9149 & 8.215 & 1.534 & .9135 & 7.719 & 1.559 & .8958 & 7.562 & 1.551 \\
& shuf           & .7736 & 10.67 & 3.273 & .7641 & 10.344 & 3.481 & .7179 & 10.155 & 3.397 \\
\hline

\hline
\end{tabular}
\end{table}

The overall picture is fully consistent with the findings of the large-scale experiments conducted on MuST-C and EuroparlST.

First, the synthetic variants of COMET exhibit very strong correlation with the standard version. However, this behavior appears to stem from the limited extent to which COMET actually exploits the source signal, as suggested by the results obtained with the shuffled source. In contrast, MetricX proves to be substantially more sensitive to both the type and the faithfulness of the source. When a reliable ASR transcript is used, the highest correlation with the standard MetricX is observed; as the WER increases, the correlation drops sharply. For the two back-translations, despite their markedly different intrinsic quality, the correlation with the standard metric remains comparable across the two conditions: lower than with the high-quality ASR, yet significantly higher than with the low-quality ASR. This pattern provides further evidence that MetricX meaningfully leverages source information, with degradation that scales with source noise in case of ASR, at a lesser extent in case of BT.

The RMSD values closely mirror the correlation results in all their nuances. In the case of COMET, they provide an even clearer indication of the impact of the different source conditions. For example, the RMSD obtained with ASRmms is more than twice that observed with ASRwhsprAda, while the RMSD values associated with the two BT variants lie roughly midway between the two ASR extremes.

Taken together, these results reinforce the robustness of the synthetic-source approach across different resource conditions, providing consistent evidence that the observed patterns are not an artifact of high-resource settings but extend to low-resource speech translation scenarios.

\subsection{Validation against Human Judgments}
\label{sec:additional-exps_DA} 

The overarching goal of this study is to assess the extent to which the synthetic, source-based variants of the two metrics can emulate their respective standard versions, as measured by correlation. This experimental design was necessitated by the lack of large-scale ST corpora providing human quality judgments for outputs produced by a diverse set of ST systems. Ideally, the validation of the synthetic variants of COMET and MetricX would have been carried out directly against such human-annotated benchmarks, had they been available. The main analyses is complemented in this section by reporting results on an ST linguistic resource, the Hearing-to-Translate test suite, that includes human quality judgments for the outputs of three speech translation systems. This additional evaluation allows to verify whether the strong correlation observed between standard and synthetic metrics is also confirmed with respect to human assessments.

\smallskip
\noindent
{\bf Correlations with the Hearing-to-Translate Human Judgements}

\noindent
Hearing to Translate~\cite{papi2025hearingtranslateeffectivenessspeech} is a test suite that provides a unified evaluation framework for assessing the effectiveness of ST systems under diverse real-world conditions. Its {\tt human evaluation} section includes annotations for automatic translations generated by a cascade composed of Canary v2 speech foundation model and Aya Expanse 32B large language model ({\tt aya\_canary}), a speech foundation model, SeamlessM4T v2 large ({\tt seamlessm}), and a SpeechLLM, Voxtral Small 24B ({\tt voxtral}). Among the 547 instances covering 10 language pairs, five of them ({\tt de-en}, {\tt en-de}, {\tt en-zh}, {\tt es-en}, {\tt it-en}) include both reference transcripts and translations, together with the three system outputs, Direct Assessment (DA) scores assigned by human annotators, and automatic scores from three Quality Estimation (QE) metrics ({\tt metricx\_qe\_normalized}, {\tt metricx\_qe\_score\_strict}, and {\tt xcomet\_qe\_by\_100}). This results in 304 fully annotated items.

Table~\ref{tab:corr-hearing2translate} reports the Pearson correlation between DA scores and the automatic scores produced by the standard and synthetic variants of COMET and MetricX, the three QE metrics included in the dataset annotations, and six additional reference-based metrics. The column {\tt avg} shows the arithmetic mean of the correlations computed over the three LLMs.

The results are noteworthy. The synthetic variants of COMET and MetricX based on {\tt ASRwhspr} correlate with DA scores even stronger than their respective standard versions. The BT-based variants exhibit slightly lower correlations, yet remain close.

When compared with the other automatic metrics, both {\tt ASRwhspr} and{\tt BTmdld} variants clearly outperform the QE metrics in terms of correlation with human judgments. With respect to the six purely reference-based metrics, the synthetic variants, like their standard counterparts, also outperform all those considered, including the corresponding versions that do not use the source, COMET-ref and MetricX-ref, although in one case the margin is negligible (COMET-BTmdld 0.5474 vs. COMET-ref 0.5471). Taken together, these results confirm two points. First, they corroborate a well-established finding in the literature: metrics that make use of the source tend to correlate better with human judgments than their counterparts that do not exploit the source. Second, they show that the synthetic metrics proposed in this work can effectively replace the standard ones while maintaining the advantage over metrics that do not use the source.

\bigskip

\begin{table}[t!]
\caption{
Pearson correlation between DA scores and scores of various evaluation metrics applied to automatic translations generated by three different ST models. Data are taken from {\tt Hearing-to-Translate} corpus. Correlations on the five language pairs are computed separately and then averaged
}
\label{tab:corr-hearing2translate}
\centering
\setlength{\dashlinedash}{0.5pt} 
\setlength{\dashlinegap}{2pt}   

\setlength{\tabcolsep}{5pt}
\begin{tabular}{ll:c|rrr:r}
\hline
\multicolumn{2}{c:}{metric} & used texts & aya\_canary & seamless & voxtral & avg \\
\hline
\multirow{3}{*}{COMET} & {\tt standard} &                       & 0.4711 & 0.5356 & 0.6581 & 0.5549 \\ 
                       & {\tt ASRwhspr} & \multirow{2}{*}{\tt src} & 0.4743 & 0.5478 & 0.6621 & 0.5614 \\
                       & {\tt BTmdld}&\multirow{2}{*}{\tt trg hyp}& 0.4655 & 0.5258 & 0.6509 & 0.5474 \\
\cdashline{1-2}\cdashline{4-7}
\multirow{3}{*}{MetricX} & {\tt standard}& 
        \multirow{2}{*}{\tt trg ref}                            & -0.4882& -0.5874& -0.6247& -0.5668 \\
                       & {\tt ASRwhspr}    &                       & -0.5097& -0.6102& -0.6375& -0.5858 \\
                       & {\tt BTmdld}     &                       & -0.4638& -0.5504& -0.6117& -0.5420 \\
\hline
\multicolumn{2}{l:}{metricx\_qe\_normalized}    & \multirow{2}{*}{\tt src ref} & 0.4464 & 0.5797 & 0.5578 & 0.5280 \\
\multicolumn{2}{l:}{metricx\_qe\_score\_strict} & \multirow{2}{*}{\tt trg hyp} & -0.4464& -0.5797& -0.5578& -0.5280 \\
\multicolumn{2}{l:}{xcomet\_qe\_by\_100}        &                              & 0.3439 & 0.5331 & 0.4675 & 0.4482 \\
\hline
\multicolumn{2}{l:}{COMET-ref}  &                             &  0.4591 &  0.5319 &  0.6502 &  0.5471 \\
\multicolumn{2}{l:}{MetricX-ref} &                            & -0.4409 & -0.4988 & -0.5898 & -0.5098 \\
\multicolumn{2}{l:}{BLEURT}     & \multirow{1}{*}{\tt trg hyp}&  0.4736 &  0.5437 &  0.6148 &  0.5440 \\
\multicolumn{2}{l:}{BLEU}       & \multirow{1}{*}{\tt trg ref}&  0.3663 &  0.4086 &  0.4714 &  0.4154 \\
\multicolumn{2}{l:}{ChrF}       &                             &  0.2965 &  0.3602 &  0.4237 &  0.3601 \\
\multicolumn{2}{l:}{TER}        &                             & -0.2984 & -0.2983 & -0.3614 & -0.3194 \\
\hline
\end{tabular}

\end{table}

\section{Discussion and Conclusions}
\label{sec:discussion}
We investigated the reliability of source-aware evaluation metrics for speech-to-text translation~(ST) when the original source transcripts are unavailable and have to be replaced with synthetic alternatives. We systematically examined two types of textual proxies, automatic transcripts (ASR) and back translations (BT), and analyzed their impact on COMET and MetricX across different datasets, languages, and models. In addition, we also delved into the problem of re-segmenting and aligning these synthetic sources with the corresponding reference translations, a critical step for computing reliable evaluation scores.

Our findings demonstrate that both synthetic sources constitute effective substitutes for manual transcripts, both under controlled conditions (Sections~\ref{sec:exp-src-controlled} and~\ref{sec:synth-source-analysis}) and in the wild (Section~\ref{sec:realistic}). This supports the use of source-aware metrics in ST evaluation (RQ1), a practice whose methodological assumptions and effects have received little attention. Between ASR-based and BT-based sources, the former exhibit superior reliability compared to the latter, provided that the transcription quality is high enough. Extensive experiments in diverse conditions indicate that case-insensitive WER, computed without punctuation, should not exceed 20\% (RQ2).\footnote{We highlight that the application of the threshold requires estimating the error rate of the ASR system used to generate the synthetic source. This information can either be inferred from known performance in similar operational conditions, a situation that rarely occurs, or measured on a suitable test set, whose availability is far from guaranteed.}
Furthermore, we showed that the proposed cross-language re-segmentation algorithm (XLR-Segmenter) allows for reliable evaluation even when audio-text alignments are unavailable (Section~\ref{sec:realistic}), yielding only negligible degradation (RQ3). The experiments also revealed that our boundary refinement step effectively restores the semantic correspondence between automatically generated ASR segments and the reference translation segments. A surprising outcome that emerged during the study is that COMET appears to make relatively limited use of the source text. Despite this, what was observed for MetricX also holds true for COMET, providing additional  support for our conclusions.

The robustness of these findings extends beyond the high-resource conditions of MuST-C and Europarl-ST. Experiments on the Bemba-English language pair (Section~\ref{sec:additional-exps_low-resource}) confirm that the observed patterns hold in a  genuinely  low-resource scenario, where both metrics and synthetic sources operate under substantially more constrained conditions. Furthermore, a direct validation against human quality judgments  (Section~\ref{sec:additional-exps_DA}) shows that the synthetic variants of COMET and MetricX reach correlations with human assessments that are even stronger than their standard counterparts, which in turn outperform all other automatic metrics considered, and are consistently superior to their source-free variants. Taken together, these results substantially reinforce the generalizability and practical relevance of our conclusions.

A natural question concerns the rationale for using BT as a textual proxy for the source audio, given that back-translations are derived from the reference translation rather than from the original speech. While this circularity may appear to undermine the added value of source-aware metrics over reference-only ones, the concern deserves a more nuanced treatment. Reference translations are themselves derived from the source audio, and therefore carry, albeit indirectly, information that originates from it. When ASR transcripts are used as source proxies, their WER quantifies the information lost with respect to the manual transcript: our results show that the retained information is sufficient for the metric to function reliably. Back-translations exhibit a lower degree of overlap with manual transcripts than ASR outputs do (as illustrated, for instance, by the WER values on the x-axis of Figure~\ref{fig:plotWER-CLvsBT} and by the BT scores in Table~\ref{tab:bembaST}). Yet this overlap is not negligible, especially considering that the semantic correspondence between BT and manual transcripts, which is what source-aware metrics actually exploit, is higher than the WER alone might suggest, as illustrated by the example in Table~\ref{tab:bonoboExample}. The empirical question is therefore whether the residual shared information is sufficient for BT-based sources to serve as effective inputs to source-aware metrics. Our results answer this question affirmatively: BT-based synthetic metrics consistently outperform their reference-only counterparts (COMET-ref and MetricX-ref) in terms of correlation with the standard metrics, and, as shown in Table~\ref{tab:corr-hearing2translate}, also in terms of correlation with human judgments.

Beyond empirical performance, we extended our analysis to practical considerations that influence the adoption of synthetic sources in real-world ST evaluation, as discussed in Section~\ref{sec:src-generation}. Regarding the \textit{language coverage}, MT systems currently offer a broader range of supported languages than ASR models, making BT a more versatile solution for low-resource or less commonly studied languages (see also Appendix~A). Concerning \textit{model quality}, our experiments show that deploying strong MT models is generally easier and more consistent (Table~\ref{tab:BT-results}), while achieving comparable ASR quality remains more challenging (Table~\ref{tab:ST-results}). When considering the \textit{neutrality of synthetic sources with respect to the ST system under evaluation}, as discussed in Section~\ref{sec:exp-src-controlled}, we observed that using ASR outputs generated by models related to the evaluated ST system introduces bias, artificially inflating metric values and reducing correlation with reference-based metrics. In this regard, BT constitutes the safest and most neutral option. The \textit{similarity between the synthetic source and the original transcripts} further influences evaluation reliability, and the results in Section~\ref{sec:exp-src-controlled} (Table~\ref{tab:winners-union-controlled}) indicate that ASR-based sources more closely approximate human transcripts, explaining their superior effectiveness when unbiased and well-aligned. Additionally, we discussed the \textit{alignment between synthetic sources and reference translations} that, unlike BT, needs to be restored for ASR sources through re-segmentation, which entails both computational overhead and a measurable drop in evaluation accuracy (Section~\ref{sec:realistic}). Therefore, BT sources should be preferred when the accuracy of the automatic alignment of ASR sources to the reference translations is critical. Lastly, in terms of overall \textit{cost} of creating synthetic sources, we carried out experiments  on a subset of data (Appendix~B). As expected, results show that the ASR generation is more demanding than BT generation.

Overall, our study provides the first systematic investigation on the deployment of source-aware metrics for ST evaluation, offering practical recommendations for their selection based on specific operating conditions. The outcomes reveal that synthetic source-aware metrics provide a reliable and effective means of evaluating ST systems, achieving strong correlation with standard metrics. By addressing both empirical and pragmatic aspects, we hope to facilitate more consistent evaluation practices in ST research.

\section{Limitations}
\label{sec:limitations}
While the findings presented in this work provide valuable insights, certain limitations should be acknowledged and are detailed below.

First, as discussed above, our approach relies on using source texts as proxies for input audio, which cannot be applied to languages without a standardized written form. These languages represent the majority of the thousands of languages spoken worldwide, and addressing this limitation would require the development of novel multimodal, source-aware metrics capable of directly leveraging the audio signal. However, this falls outside the scope of this work.

A second limitation concerns language coverage. The two main datasets used in our study cover only high- and medium-resource languages, which reflects an intrinsic limitation of modern neural evaluation metrics that, being data-driven, generally perform less reliably on low-resource languages. Our experimental setup was therefore designed consistently with the current capabilities of state-of-the-art metrics. The experiments on Bemba–English (Section~\ref{sec:additional-exps_low-resource}) extend our investigation to a low-resource scenario and confirm the robustness of our findings in that setting. However, this single language pair cannot be considered exhaustive with respect to the full spectrum of challenges posed by low-resource conditions. In particular, our study does not cover language pairs in which both languages are low-resource, nor scenarios where no pretrained ASR or MT models are available for the languages involved, cases in which the synthetic source generation strategies proposed in this work may not be directly applicable.

A related limitation concerns the scale of the human judgment validation. The Hearing-to-Translate test suite used in Section~\ref{sec:additional-exps_DA} comprises only 304 fully annotated items, covering five language pairs and three ST systems. While the results are positive and consistent with the findings of the main experiments, the limited size of this dataset means that the conclusions drawn from this validation should be interpreted with appropriate caution. A more definitive assessment of the correlation between synthetic metrics and human judgments would require larger-scale human annotation efforts, covering a broader range of language pairs, translation systems, and domains. As with the Bemba–English experiments, the evidence is compelling but cannot be taken as conclusive across all possible evaluation scenarios.

A further limitation relates to the nature of the benchmarks used. All datasets consist of clean, single-speaker recordings without background noise, rather than speech recorded in natural conditions. While this setting may not capture the full complexity of real-world scenarios, these datasets remain the only benchmarks currently available for speech translation that cover such a broad range of languages (i.e., 79 language pairs).

Another consideration is that our analysis focuses on only two source-aware evaluation metrics, COMET and MetricX. Since MetricX is the only metric significantly affected by the source, some detailed results for COMET were reported in Appendix~E to enhance clarity and focus of the paper. Although this scope may appear narrow, these metrics were deliberately selected as widely used representatives of two distinct families of source-aware metrics, both consistently ranked among the top-performing ones in the WMT Metrics Shared Tasks. Furthermore, the fact that the findings obtained with MetricX remain valid for COMET further supports the broader applicability and reliability of our conclusions.

A final limitation concerns the quality of the systems used to generate synthetic sources. The two MT models employed for back-translation are both high performing, whereas the three ASR systems used to produce source ASRs do not always reach the same level of accuracy. We acknowledge that including lower-performing MT models in the analysis would likely have resulted in a reduction of the reliability of BT sources, similar to the degradation observed for ASR sources with high WERs. However, in typical research and experimental conditions, high-quality MT models are generally easier to obtain than equally reliable ASR systems, as discussed in Appendix A. Accordingly, our experimental setup was designed to be consistent with the capabilities of systems commonly available in current research environments. Again, the experiments on Bemba-English (Section~\ref{sec:additional-exps_low-resource}) partially address this limitation, as they include BT models of markedly different quality, even rather weak, and show that BT-based synthetic metrics retain their effectiveness even when the back-translation quality is low. However, as with the limitation on language coverage, this evidence is drawn from a single language pair and cannot be regarded as a general guarantee, particularly for scenarios involving language pairs for which only very low-quality MT models are available.

\appendix

\appendixsection{On the Quality of ASR and MT Models}
\label{app:quality}
In Section~\ref{sec:src-generation}, we discussed that MT models tend to exhibit higher quality than ASR models. In the following, we outline typical performance levels reported in the literature for both tasks, to contextualize our findings and clarify the expected reliability of the synthetic sources used in our experiments.

For ASR, transcription quality is influenced by a wide variety of factors, leading to outcomes that range from near-human accuracy to levels of limited practical utility. For clean, read speech (e.g., the {\tt test-clean} split of LibriSpeech), current systems achieve WERs of 1-3\%~\cite{Gulati2020ConformerCT,wav2vecBaevskiEtAl2020}.
On spontaneous speech (e.g., Switchboard, GigaSpeech), WER typically rises to 5-10\% or higher~\cite{conformerZeineldeenEtAl2022,ng21b_interspeech}, as also reflected on the Hugging Face Open ASR leaderboard~\cite{open-asr-leaderboard}. 
For instance, the Canary-1B model reports an average WER of 6.5\%, less than one point above the best score of 5.6\%,\footnote{Visited on 29 Aug 2025} but its performance on spontaneous telephone conversations from the CallHome benchmark drops to 16.9\% WER~\cite{canaryWangEtAl2024}, showing the substantial degradation under less controlled conditions.

More challenging operating conditions are proposed annually by the CHiME challenge. In the 2024 edition,\footnote{\url{https://www.chimechallenge.org/workshops/chime2024/}} two ASR tasks were organized: DASR (multi-channel distant ASR) and NOTSOFAR (single-device meeting transcription). Both tasks combine spontaneous speech with multiple (sometimes overlapping) speakers and variable microphone distances, resulting in WERs ranging from 20\% to 50\%.\footnote{\url{https://huggingface.co/spaces/NOTSOFAR/CHiME8Challenge}}
Transcribing non-native speech or speech with strong regional accents is also similarly challenging: \cite{accentedAsrDoEtAl2024} reports WERs above 30\% on a test set of 40 English accents, despite the model achieves 1-3\% WER on clean, read speech.
Finally, dysarthric speech represents another critical condition~\cite{dysarthricQianEtAl2023}, with transcription accuracy highly varying with the severity of dysarthria and exceeding 60\% WER with high levels of impairment~\cite{dysarthricAlmadhorEtAl2023}.

For MT, as for ASR, performance scores vary substantially depending on domain, language pair, and, most importantly, the amount of available training data. Notably, the OPUS-MT Dashboard\footnote{\url{https://opus.nlpl.eu/dashboard/}, visited on 29 Aug 2025} shows that high-resource language pairs achieve consistently high performance not only for closely related languages, such as English-French or English-German for which BLEU scores often exceed 40 and COMET scores 85-90, but also for linguistically distant ones, such as English-Chinese or English-Japanese. In contrast, substantial drops in quality are observed mainly for very low-resource languages, such as Yoruba or Wolof, where the scarcity of training data imposes severe constraints on model performance.

In conclusion, if ASR can be problematic due to the intrinsic difficulties of the task even in quite common conditions (meetings, non-native speakers), the main problem for MT does not lie in the task itself but in the availability of training data.  This supports our choice to conduct the investigation including also rather poor ASR models but only good quality MT models.

\appendixsection{Cost of Generating Synthetic Sources}
\label{app:cost}
To estimate the computational cost associated with generating the two types of synthetic sources, we executed each step of the respective pipelines independently on a single Tesla A40 GPU with 48GB of memory. For each step, we measured the execution time of its core operations (excluding overheads such as model loading), and recorded the peak GPU memory usage. The evaluation was conducted on the es$\rightarrow$it Europarl-ST test set, which contains approximately 3 hours and 8 minutes of Spanish audio. 

\begin{table}[h!]
\caption{Costs in terms of total execution time (excluding the model loading) and GPU memory usage of the various steps required for generating either ASR or BT  sources. The size of models in terms of number of parameters is also shown}

\label{tab:costs}
\setlength{\tabcolsep}{4pt}
\centering
\begin{tabular}{lllcc|cc}
\hline
id & step & model & size & batch & time & GPU  \\
&          &       &      & size  &      & memory \\
\hline
\multirow{4}{*}{1} & \multirow{4}{*}{MT}  & \multirow{2}{*}{MADLAD} & \multirow{2}{*}{3B}   & 1  & 20m:30s & 12.5GB \\
 &    &        &      & 32 &  2m:48s & 44.6GB \\
    \cdashline{3-7}
 &   & \multirow{2}{*}{NLLB}   & \multirow{2}{*}{3.3B} & 1  & 12m:47s & 13.7GB \\ 
 &   &        &      & 32 &  3m:02s & 42.6GB \\
\hdashline
\multirow{3}{*}{2} & \multirow{3}{*}{ASR} & Whisper & \multirow{1}{*}{1.55B} & 1 & 21m:26s & 9.7GB \\ 
\cdashline{3-7}
 &     & \multirow{2}{*}{SeamlessM4T} & \multirow{2}{*}{2.3B} & 1 & 17m:45s & 10.8GB \\ 
 & & & & 32 & 3m:43s & 38.7GB \\ 
\hdashline
3 &  L-Segmenter & mweralign & - & 1 & 20s & - \\
4.a & \multirow{2}{*}{\underline{R}efinement} & SimAlign (mBERT) & 180M & 1 & 21m:35s & 1.1GB \\
4.b & & LaBSE & 500M & 1 & 2m18s & 2.6GB \\
\hline
\end{tabular}
\end{table}

The resulting values, reported in Table~\ref{tab:costs}, allow to compare the resource demands of the ASR- and BT-based pipelines: since the generation of BT sources requires only the MT step, while the generation of ASR sources requires all four steps, the BT pipeline is at least 2-8 times faster than the ASR pipeline, depending on the re-segmentation type (XL-Resegmenter, XLR-SimAlign/LaBSE) and on the level of hardware parallelism exploitation within the GPU via larger batch sizes. One of the factors driving this significant difference is the high cost of the segment boundary refinement Algorithm~\ref{alg:refinement} (Table~\ref{tab:costs}, id 4), especially when SimAlign is used. It is important to note that Algorithm 1 is not  parallelizable by design, since shifting the right boundary of a segment changes the left boundary of the next segment (steps 3-5), preventing their concurrent processing. Ultimately, given the same available hardware and parallelism exploitation, BT sources are significantly more cost-effective to generate than ASR sources. It is worth noting that, should one opt for ASR over BT sources, the choice of re-segmentation strategy involves a trade-off between accuracy and efficiency. XL-Segmenter alone is by far the most efficient option, but also the least accurate, as shown by the results presented in the main paper. The refinement stage substantially improves accuracy: if maximum accuracy is the priority, the SimAlign-based variant is the preferred choice; however, accepting only a minimal accuracy degradation, LaBSE may be a preferable option given its significantly lower latency.

\begin{figure}[h!]
\begin{center}
\includegraphics[width=1.0\textwidth]{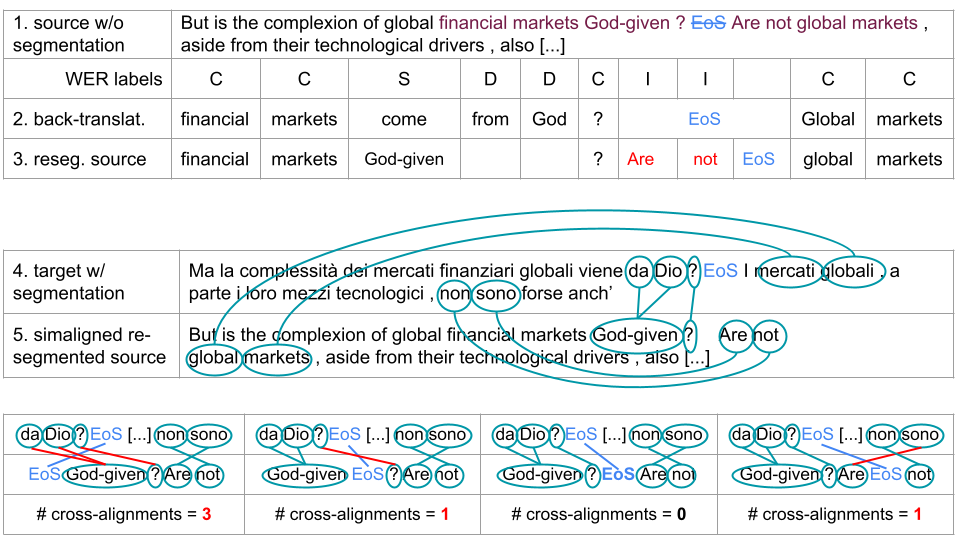}
\end{center}
\caption{Operation of the two-stage re-segmentation algorithm on a real example under controlled conditions (re-segmentation of manual transcripts).}
\label{fig:resegExample}
\end{figure}

\appendixsection{The Two-Stage Re-Segmentation Algorithm at Work}
\label{app:resegm}
Figure \ref{fig:resegExample} illustrates how the two-stage re-segmentation algorithm works on a controlled example, where the objective is to realign the reference transcript, whose gold segmentation is disregarded, with the segments of the reference translation. The results of experiments conducted in this controlled mode are reported in Section~\ref{sec:reseg-controlled}.

Row 1 of the table contains the reference transcript, with the portion of text under focus highlighted in {\color{RedViolet} red violet}; the original segmentation point is also shown but ignored ({\color{RoyalBlue}\sout{EoS}}), in order to evaluate the algorithm's ability to correctly (re)identify it. Row 4 contains the segmented reference translation, while the back-translation of its segments is shown in Row 2. The contents of rows 1 and 2 represent the input to L-Segmenter. It outputs the re-segmentation of the row 1 text as shown in row 3, derived from the ``WER labels'' assigned to \underline{C}orrect word matches, \underline{S}ubstitutions, \underline{I}nsertions, and \underline{D}eletions. Notably, near the {\color{RoyalBlue}EoS} marker, whose placement derives from its position in the back-translation, the re-segmented source has two consecutive words, ``\textcolor{red}{Are not}'', that are not aligned to any word in the back-translation and are therefore labeled as  \underline{I}nsertions. These words could, in principle, be assigned to the left segment, to the right segment, or split between them. In the example shown, both words are (\textcolor{red}{incorrectly}) assigned to the left segment.

The boundary refinement stage is then applied, aligning words using the SimAlign algorithm. Working at the embedding level, SimAlign can identify correspondences not only between words in different languages but also in a more robust manner than the Levenshtein distance minimization does. Rows 4 and 5  (partially) show the word alignments generated by SimAlign between the reference translation and the text obtained by concatenating the two segments of row 3 (Step 5 of Algorithm~\ref{alg:refinement}). Given this alignment, the number of cross-alignments (see the lower table in the figure) is minimized (\#cross-alignments=0) precisely when the moving {\color{RoyalBlue}EoS} tag is placed to ({\color{RoyalBlue}{\bf correctly}}) reassign the two words ``Are not'' to the right segment.

\appendixsection{Systems' Ranking Correlation}
\label{app:ranking}
As further confirmation of the reliability of our synthetic variants of COMET and MetricX, we verified the preservation of the ranking among the ST systems by calculating the correlations between the system scores on the entire test sets. For each synthetic variant (based on either ASR- or BT-generated sources) of each source-aware metric,  and for every test set and language pair, we computed the correlation between the scores of the six ST systems considered in our experiments as measured by the synthetic and by the corresponding standard version of the metric. The values shown in Table~\ref{table:sys_corr} are averages of these correlations across all language pairs in the two datasets, and are reported both for the controlled setting, where the alignment between reference translations and source audio is known, and for the uncontrolled setting, where it is not. As we can see, the correlation is always higher than 0.99, with the only exception of OWSM with known audio-reference alignment on MuST-C. However, correlations are very high for all methods. This enforces that not only systems' rankings are stable regardless of the source and condition, but also the differences between the scores obtained by different systems are not significantly impacted by the chosen method.

\begin{table}[ht!]
\caption{COMET/MetricX correlations of systems' scores averaged across language pairs in each test set}
\label{table:sys_corr}
\centering
\begin{tabular}{l|cc|cc}
\toprule
  synthetic & \multicolumn{2}{c|}{COMET}  & \multicolumn{2}{c}{MetricX} \\
  source    & EP-ST & MuST-C & EP-ST & MuST-C \\
\midrule
{\tt ASRwhspr} & 0.9993 & 0.9992 & >0.9999 & 0.9999 \\
{\tt ASRowsm} & 0.9925 & 0.9879 & 0.9999 & 0.9986 \\
{\tt ASRsmlss} & 0.9995 & 0.9922 & >0.9999 & 0.9980 \\
\hdashline
{\tt BTmdld} & 0.9954 & 0.9999 & >0.9999 & >0.9999 \\
{\tt BTnllb} & 0.9948 & 0.9999 & >0.9999 & >0.9999 \\
\hline
{\tt ASRwhspr-mdld} & 0.9999 & 0.9999 & >0.9999 & >0.9999 \\
{\tt ASRwhspr-nllb} & 0.9999 & 0.9999 & >0.9999 & >0.9999 \\
{\tt ASRowsm-mdld} & 0.9977 & 0.9992 & 0.9999 & 0.9999 \\
{\tt ASRowsm-nllb} & 0.9977 & 0.9992 & 0.9999 & 0.9999 \\
{\tt ASRsmlss-mdld} & >0.9999 & >0.9999 & >0.9999 & 0.9999 \\
{\tt ASRsmlss-nllb} & >0.9999 & >0.9999 & >0.9999 & 0.9999 \\
\bottomrule
\end{tabular}
\end{table}

\appendixsection{COMET in the Wild (Realistic Conditions)}
\label{app:comet}
Tables~\ref{tab:comet-corr-mustc-shas} and~\ref{tab:comet-corr-epst-shas} show for COMET the scores that Tables~\ref{tab:corr-mustc-shas} and~\ref{tab:corr-epst-shas} show for MetricX. They provide, for the two corpora, the correlations related to the various ASR  sources generated in realistic conditions, and the corresponding gap recovery with respect to the shuffled sources. The ASR sources are those available at the end of the two-stage re-segmentation algorithm, whose quality is provided in rows labeled as {\tt shas+XLR-Segmenter} of Table~\ref{tab:in-vivoReseg}. Values have to be compared to those of Comet$_{\tt BTmdld}$ and Comet$_{\tt BTnllb}$ in Tables~\ref{tab:corr-mustc} and~\ref{tab:corr-epst}, which are shown in Tables ~\ref{tab:comet-corr-mustc-shas} and~\ref{tab:comet-corr-epst-shas} as well for ease of reading.

\begin{table}[h!]
\caption{Pearson correlations between the COMET scores of each system on MuST-C, computed (i) by using as the source the reference transcript (i.e., in the standard way), and (ii) by using as the source the automatically re-segmented automatic transcripts generated from automatic audio segmentation. Each reported correlation coefficient is the average of the correlation coefficients computed separately for each language pair. For \textdagger~(correlations potentially biased) and heat map colouring see caption of Table~\ref{tab:corr-mustc}}

\label{tab:comet-corr-mustc-shas}
\setlength{\tabcolsep}{2pt}
\includegraphics[width=1.0\textwidth]{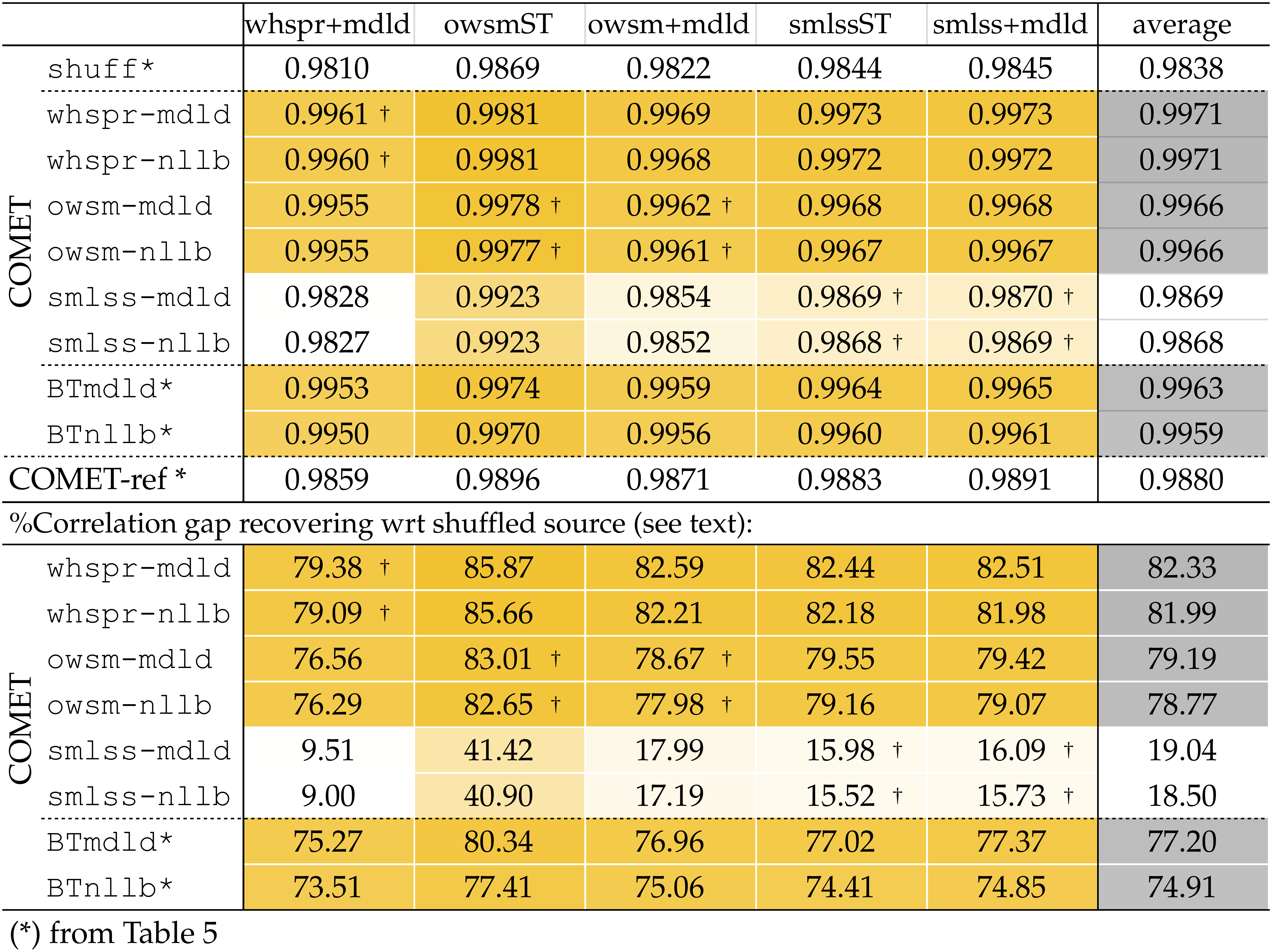}
\end{table}

\begin{table}[ht]
\caption{Pearson correlations between the COMET scores of each system on Europarl-ST, computed in the standard way, i.e. by using as the source the reference transcript, and the COMET scores computed using as the source the automatically re-segmented automatic transcripts generated from automatic audio segmentation. Each reported correlation coefficient is the average of the correlation coefficients computed separately for each language pair.  For \textdagger~(correlations potentially biased) and heat map colouring see caption of Table~\ref{tab:corr-mustc}}

\label{tab:comet-corr-epst-shas}
\setlength{\tabcolsep}{2pt}

\includegraphics[width=1.0\textwidth]{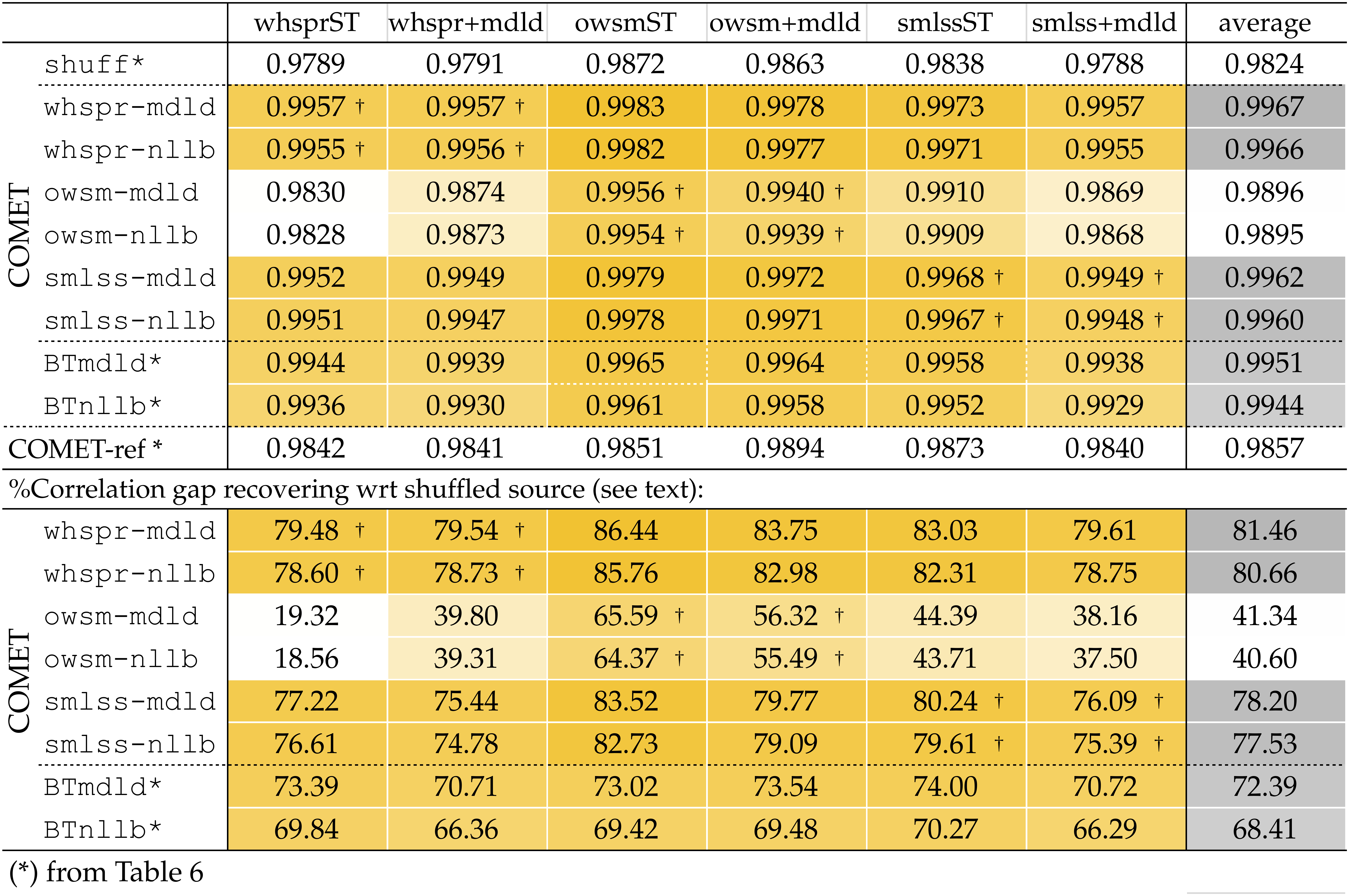}
\end{table}

\begin{figure}[ht]
\begin{center}
\includegraphics[width=0.5\textwidth]{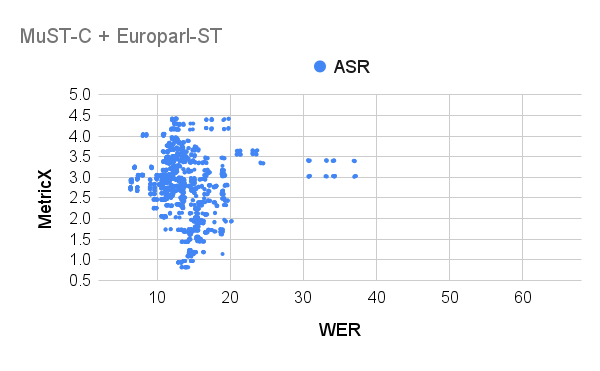}\includegraphics[width=0.5\textwidth]{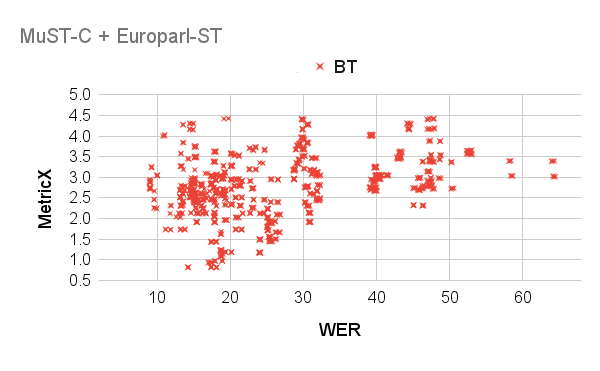}
\end{center}
\caption{
The plane of these two scatter charts is defined by the WER and MetricX scores of the ASR and BT sources, respectively. Considering all unbiased pairwise combinations between language pairs in each corpus and ST systems, they show where the cases in which it was preferable to use the ASR (on the left) or the BT (on the right) as a source for the computation of COMET are placed in that plane. The total number of points is 3440, 2185  on the left (ASR wins, 63.5\%), 1255 on the right (BT wins, 36.5\%). A random 1\% change was applied to all values in order to avoid the overlapping of points and make all of them visible.} 
\label{fig:comet-plotUnion-realistic}
\end{figure}

\begin{figure}[ht]
\begin{center}
\includegraphics[width=0.5\textwidth]{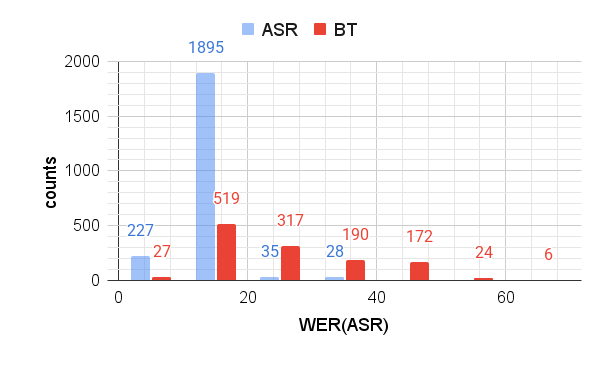}\includegraphics[width=0.5\textwidth]{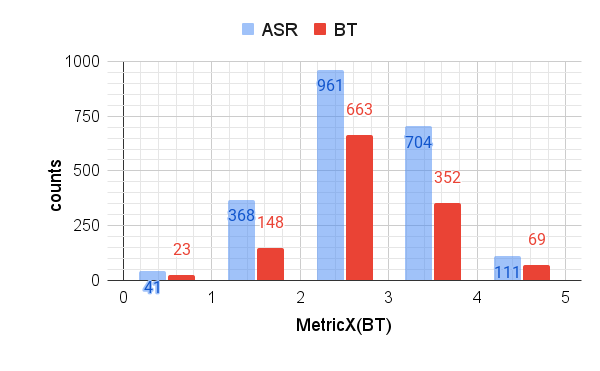}
\end{center}
\caption{Considering all unbiased pairwise combinations of language pairs in each corpus and ST systems, these histograms illustrate the distribution of cases in which the standard COMET shows a higher correlation with COMET using either the ASR or the BT as source input. The left plot reflects this distribution as a function of transcription quality (WER), while the right plot does so with respect to (back-)translation quality (MetricX). }
\label{fig:comet-histoUnion-realistic}
\end{figure}

Plots in Figure~\ref{fig:comet-plotUnion-realistic}, histograms in Figure~\ref{fig:comet-histoUnion-realistic}, and Table~\ref{tab:comet-winners-union-realistic} are equivalent to Figures~\ref{fig:plotUnion-realistic},~\ref{fig:histoUnion-realistic} and Table\ref{tab:winners-union-realistic}, respectively,  related to COMET metric instead of MetricX.

Despite the weak dependence of COMET on the source, the empirical evidences observed for MetricX remain valid for COMET as well; among these, there is the 20\% WER threshold that distinguishes whether the ASR source quality is sufficient or not to make it preferable to the BT source.

\begin{table}[ht]
\caption{
For all possible comparisons between the COMET correlation with the ASR source and that with the BT source, computed on all language pairs of the two corpora and for all ST systems, the total number of wins per synthetic source type (ASR - whisper, owsm, seamless resegmented by XLR-Segmenter wrt madlad or nllb - or BT - madlad, nllb) is given here. Biased ASR COMETs, i.e., those of ST systems that are somehow involved in the generation of the ASR source of the metric, are excluded}

\label{tab:comet-winners-union-realistic}
\setlength{\tabcolsep}{1.7pt}
\centering
\begin{tabular}{l|cccc|cccc|cccc}
\hline
\multirow{3}{*}{corpus} 
& \multicolumn{4}{c|}{WER$_{\tt ASR}\le$20\%} & \multicolumn{4}{c|}{WER$_{\tt ASR}>$20\%} & \multicolumn{4}{c}{total} \\
& \multicolumn{2}{c}{ASR wins} & \multicolumn{2}{c|}{BT wins}& \multicolumn{2}{c}{ASR wins} & \multicolumn{2}{c|}{BT wins}& \multicolumn{2}{c}{ASR wins} & \multicolumn{2}{c}{BT wins} \\
& counts & \% & counts & \% & counts & \% & counts & \% & counts & \% & counts & \% \\
\hline
MuST-C       & 300  &  89.3 & 36  & 10.7 & 56  & 25.9 & 168 & 75.0 & 356  & 63.6 & 204  & 36.4 \\
Europarl-ST  & 1822 &  78.1 & 510 & 21.9 &  7  &  1.3 & 541 & 98.7 & 1829 & 63.5 & 1051 & 36.5 \\
\hline
total        & 2122 &  79.5 & 546 & 20.5 & 63  &  8.2 & 709 & 91.8 & 2185 & 63.5 & 1255 & 36.5 \\
\hline
\end{tabular}
\end{table}


\clearpage

\bibliographystyle{compling}
\bibliography{biblio}

\end{document}